\documentclass[11pt]{article}

\usepackage[utf8]{inputenc}
\usepackage{style}

\usepackage[T1]{fontenc}    %
\usepackage{url}            %
\usepackage{booktabs}       %
\usepackage{amsfonts}       %
\usepackage{nicefrac}       %
\usepackage{microtype}      %
\usepackage{xcolor} %
\usepackage{float}

\usepackage{xcolor}
\usepackage{mathtools,amsfonts,amssymb,mdframed,xspace,xfrac,multicol,bbm,bm}
\usepackage{makecell}
\usepackage{eucal}
\usepackage{amssymb,amsmath,amsopn,mathtools}
\usepackage{mathrsfs}
\usepackage{complexity}
\usepackage{enumitem}
\usepackage{subcaption}
\usepackage{caption}
\usepackage{dsfont}
\usepackage{wrapfig}
\usepackage{framed}
\usepackage[capitalize,noabbrev]{cleveref}
\usepackage{diagbox}
\usepackage{natbib}

\newcommand{\eps}{\varepsilon}

\newcommand{\Ex}{\mathbb{E}}
\renewcommand{\E}{\mathbb{E}}

\newcommand{\Var}{\operatorname{Var}}

\renewcommand{\opt}{{\sf{opt}}}
\renewcommand{\val}{{\sf{val}}}
\renewcommand{\R}{\mathbb{R}}

\newcommand{\LFA}{{{\sc wtd-FLAg}\,}}
\newcommand{\LBA}{{{\sc sampled-wtd-FLAg}\,}}
\newcommand{\LLP}{{{\sc sampled-noisy-wtd-LAg}\,}}

\newcommand{\mc}[1]{\ensuremath{\mathcal{#1}}\xspace}
\newcommand{\mb}[1]{\ensuremath{\mathbf{#1}}\xspace}
\newcommand{\tn}[1]{\ensuremath{\textnormal{#1}}\xspace}
\newcommand{\ol}[1]{\ensuremath{\overline{#1}}\xspace}

\newcommand{\bX}{{\mb{X}}}
\newcommand{\bY}{{\mb{Y}}}

\newcommand{\bZ}{{\mb{Z}}}
\newcommand{\ba}{{\mb{a}}}
\newcommand{\bb}{{\mb{b}}}

\newcommand{\bW}{{\mb{W}}}
\newcommand{\bU}{{\mb{U}}}
\newcommand{\bV}{{\mb{V}}}

\newcommand{\bv}{{\mathbf{v}}}
\newcommand{\bw}{{\mathbf{w}}}
\newcommand{\br}{{\mathbf{r}}}

\newcommand{\bD}{{\mathbf{D}}}

\title{Label Differential Privacy via Aggregation}
 \author{Anand Brahmbhatt* \\
 \and Rishi Saket*\\
 \and Shreyas Havaldar*\\
 \and Anshul Nasery*\\
 \and Yukti Makhija*\\
 \and Aravindan Raghuveer*}
 
\newcommand\blfootnote[1]{%
  \begingroup
  \renewcommand\thefootnote{}\footnote{#1}%
  \addtocounter{footnote}{-1}%
  \endgroup
}

\begin{document}
\date{}
\maketitle
\blfootnote{* Google Research India. \texttt{\{anandpareshb,rishisaket,shreyasjh,anshulnasery,araghuveer\}@google.com}}

\begin{abstract}

This paper explores the use of linear aggregation to protect the privacy of sensitive training labels through the concept of \emph{label differential privacy} (label-DP) while maintaining regression task utility. Our key finding is that weighted linear aggregation of training instances with i.i.d. $N(0, 1)$ weights can achieve $(\eps, \delta)$-label-DP with 
$m = O\left(n/(\log(1/\delta))\right)$. Unlike prior methods, our approach relies on the minimum linear regression loss rather than the minimum singular value of the data matrix, resulting in better practical bounds on real datasets.

We also examine real-world mechanisms involving disjoint sets or \textit{bags} of instances. We demonstrate that aggregating labels from sub-sampled disjoint $k$-sized bags using i.i.d. $N(0,1)$ weights achieves $(\eps,\delta)$-label-DP with $k \geq \Omega\left(\left((1/\eps)\log\left(1/\delta\right)\right)^2\right)$. In both scenarios, the optimal linear mse-regressor on the aggregated data approximates the original dataset's optimum with high probability, without needing additive label noise.

Furthermore, we show that adding $N(0,1)$ noise to any constant fraction of labels allows for similar label-DP guarantees when aggregating labels over random disjoint bags, while preserving the utility of Lipschitz-bounded neural mse-regression tasks.
\end{abstract}

\section{Introduction}
The notion of \emph{differential privacy}, introduced by \citep{dwork2006differential}, has become a widely studied measure of privacy guarantees for randomized \emph{mechanisms}, i.e., data processing algorithms. Roughly speaking, it states that the output distribution of a mechanism should not change significantly with \emph{unit} changes to the input data. Thus, any adversarial modification to a small part of the input data is unlikely to substantially impact the mechanism's output. For example, in machine learning, a differentially private (DP) model trainer would be robust to changes in a single training example, i.e., a pair of feature vector and its label.

In many cases, the sensitive information may entirely reside in the labels of the training feature vectors, such as the votes of electors or the outcomes of medical tests. The corresponding mechanisms have been studied using the notion of \emph{label differential privacy} (label-DP) \citep{chaudhuri2011sample}, in which the input data changes only in one training label. In DP mechanisms, the guaranteed bound on the distributional shift is typically achieved by adding independent noise to each element (or label, in the case of label-DP) of the dataset.

Recently, aggregation-based frameworks, such as \emph{learning from label proportions} (LLP), have gained importance due to privacy concerns \citep{R10,o2022challenges}. In LLP, the training data is partitioned into \emph{bags}; while the constituent feature vectors of each bag are available, only the sum of the labels in the bag can be observed, thereby obfuscating the individual labels. We refer to mechanisms in which only the labels in each bag are aggregated as \emph{label aggregation} (LAg). In some situations, due to instrumentation or efficiency considerations \citep{WIBB}, both the labels and feature vectors in bags are linearly aggregated in what we call \emph{feature and label aggregation} (FLAg) mechanisms, with the learning problem being \emph{learning from bag aggregates} (LBA). Unlike additive noise-based DP mechanisms, aggregation mechanisms like LAg and FLAg naturally occur in various practical applications, especially due to systemic or recent regulatory limitations (e.g., Apple SKAN \citep{Skan} and Chrome Privacy Sandbox \citep{priv}). We believe it is important to understand the privacy guarantees provided by such mechanisms. This may help, for e.g. in deciding whether or to what extent additional privacy preserving mechanisms are required, potentially saving unnecessary investments for implementing them. Towards this, we study label-DP guarantees for LAg and FLAg mechanisms, either without additive noise or in conjunction with (more efficient) noise mechanisms, while simultaneously preserving the utility of natural regression tasks.

It is easy to observe that unweighted summation-based LAg and FLAg are not label-DP on their own when the dataset is randomly partitioned into one or more bags (see Appendix \ref{sec:no-DP-example}). So the question then becomes whether and in what scenarios can enhancements to LAg and FLAg provide label-DP guarantees while providing the desired utility bounds.

{\bf Informal description of our results.} We propose a generic bag sampling process with i.i.d. weights for the instances in each bag. Using this we define a \emph{weighted} version of FLAg as follows: given a collection of $m$ subsets (bags), the mechanism outputs for each bag the weighted sum of the feature-vectors as the aggregate feature-vector along with the weighted sum of the labels as the aggregate label. We consider two variants depending on how the bags are constructed: weighted-FLAg in which each of the $m$ bags is the entire set of feature-vectors in the dataset, and sampled-weighted-FLAg in which the $m$ bags are a random collection of $k$-sized disjoint bags. Note that no \emph{additive} noise is added to the labels.

Our work shows that label-DP can be achieved using i.i.d. Gaussian weights via the  weighted-FLAg mechanism as well as sampled-weighted-FLAg under mild assumptions on the non-degeneracy of the dataset labels, while preserving the utility (optimum) of linear regression using bounded norm regressors with the \emph{mean squared error} (mse) loss. The advantage of sampled-weighted-FLAg is that it requires aggregating a much smaller portion of the dataset, the drawback being that the bounds also depend on linear non-degeneracy of the features, given by a spectral parameter.

In the case of LAg, the bag feature-vectors along with their weights are individually made available. To overcome the consequent loss in privacy, the mechanism does addition of i.i.d. Gaussian noise of \emph{unit variance} to only a \emph{fixed small fraction} of the labels followed by weighted label aggregation with Gaussian weights to ensure label-DP. We call this the sampled-noisy-weighted LAg which we show preserves the utility w.r.t. the noisy dataset of neural mse-regression using neural networks with Lipschitz-bounded outputs and bounded weights, which includes the linear regression task mentioned above.

\textbf{Contributions of this paper:} The primary motivation of this work is to quantify the privacy achieved through aggregation methods.
\begin{itemize}[leftmargin=*]
    \item We establish that merely aggregating labels, as done in the construction of LLP datasets, does not provide differential privacy guarantees.
    \item We introduce novel aggregation algorithms that take a dataset as input and produce a modified label-differentially private dataset \emph{solely through aggregation, without using additive noise}.
    \item We demonstrate that the datasets generated by these label-DP algorithms preserve utility for linear regression tasks.
\end{itemize}

{\bf Organization of the paper.} Section \ref{sec:related-work} provides an overview of the previous work on aggregation mechanisms and techniques for model training on aggregated data. Section \ref{sec:Ourresults} contains the descriptions of the weighted-FLAg and sampled-weighted-FLAg mechanisms along with our results on their privacy and utility preservation. The same for sampled-noisy-weighted LAg is deferred to Appendix \ref{sec:LLP-mechanism-results}. Section \ref{sec:discussion} places our results in context of related work mentioned above, while Section \ref{sec:overview} provides an informal description of the proof techniques. Section \ref{sec:privacy_LFA} provides the privacy analysis for weighted-FLAg and that for sampled-weighted-FLAg is given in Section \ref{sec:privacynet}. An empirical evaluation of the privacy guarantees as well as the utility afforded by  weighted-FLAg is provided in Section \ref{sec:experiments}.

\section{Related Work}\label{sec:related-work}

\textbf{Learning from aggregated data.}
The LLP problem has been studied in the context of label privacy concerns~\citep{R10}, costly supervision~\citep{CHR} or lack of labeling instrumentation~\citep{DNRS}. Recent restrictions on user data on advertising platforms -- due to aggregate conversion label reporting systems (\citep{Skan,priv}) --  have led to further interest in model training on aggregated labels~\citep{o2022challenges}. On the other hand, the LBA problem occur where (e.g. in \citep{WIBB}) the individual feature-vectors as well as labels are sensitive.  
LLP was studied in  \citep{FK05,HIL13,MCO07,R10} which adapted supervised learning techniques for it, while  \cite{QSCL09} and \cite{PNCR14} obtained algorithms based on bag-label mean estimates and \cite{YLKJC13} proposed a novel SVM with bag-level constraints. Subsequently, methods using bag pre-processing \citep{SZ20,ZWS22,SRR}, deep learning%
~\citep{KDFS15,DZCBV19,LWQTS19,NSJCRR22} as well as belief propagation for pseudo-labeling in the LLP setup~\citep{havaldar2023learning} have been developed. 
Recent works have proposed techniques for either randomly sampled~\citep{busafekete2023easy} or curated bags~\citep{chen2023learning}. The theoretical aspects of LLP were studied by \citep{YCKJC14} who defined it in the PAC framework while \citep{Saket21,Saket22} have shown worst case algorithmic and hardness bounds, for the bag-classification task using linear classifiers. On the other hand, recent work~\citep{brahmbhatt2023pac} showed accurate PAC learning algorithms for Gaussian feature vectors and random bags.

\textbf{Label Differential Privacy}. 
Differential privacy was introduced in \citep{dwork2006differential} and it has become a popular notion of privacy. Subsequently, \cite{dwork2014algorithmic} provided a detailed study of methods to develop DP algorithms. %
The work of \cite{chaudhuri2011sample} first introduced the restricted notion of label-DP and proved a lower bound on the excess risk of learning with such guarantees. Recently, \cite{ghazi2021deep} provided an algorithm to achieve label-DP for classification by adding noise to the labels, which was extended by \cite{ghazi2022regression} to regression tasks, while \cite{malek2021antipodes} incorporated Bayesian inference to improve the utility of \cite{ghazi2021deep}, and \cite{esfandiari2022label} proposed a label-DP mechanism based on clustering the feature-vectors and resampling their labels.%

\textbf{Differentially Private Linear Regression.} Since linear regression is widely applied on real data, achieving this task with DP guarantees has been the focus of several previous works such as \citep{Sheffet-OLS,Sheffet-old,Cai,Wang-2,Alabi,Amin} which proposed DP algorithms for linear regression and its variants. Some of these works, e.g. \cite{Sheffet-OLS}, \cite{Sheffet-old} and \cite{Wang-2}, provide mechanisms which use weighted aggregation to obfuscate the input dataset. Most relevant to our work is that of \cite{Sheffet-old} which analyzes the Johnson-Lindenstrauss random projection. While the resultant aggregation is same as the weighted-FLAg aggregation studied in this work, instead of label-DP, \cite{Sheffet-old} proves DP guarantees under the stronger non-degeneracy assumptions.%

\section{Problem Definition, Our Aggregation Mechanisms and Results}\label{sec:Ourresults}
A dataset $\bD = \{(\bx_{i}, y_{i})\}_{i=1}^n$ consists of labeled $(d-1)$-dimensional feature vectors. For any class $\mc{C}$ of real-valued regressors over $\mathbb{R}^d$, the mean squared error (mse) regression task is to compute $f_* \in \mc{C}$ that minimizes $\val(\bD, f) := \sum_{i=1}^n \left(y_{i} - f\left(\bx_{i}\right)\right)^2$. We define $\opt(\bD, \mc{C}) = \min_{f \in \mc{C}} \val(\bD, f)$.

When the class $\mc{C}$ contains only linear functions, we absorb the constant term into the feature vectors by appending a $1$-valued coordinate to each vector. We define $(\bX, \by)$ as the matrix representation of dataset $\bD$ where $\bX = [\bx_{i}^{\sf T}]_{i=1}^n \in \mathbb{R}^{n \times d}$, and $\by = [y_{i}]_{i=1}^n \in \mathbb{R}^{n \times 1}$. Notice that $\val(\bD, \br^{\sf T}\bx) = \|\by - \bX\br\|_2^2$. We denote $\mc{F}$ as the class of all homogeneous linear regressors and $\mc{F}_{B_r}$ as the subclass of $\mc{F}$ such that for any $f \in \mc{F}$, $f(\bx) := \br^{\sf T}\bx$ where $\|\br\|_2 \leq B_{\sf r}$ for some $B_{\sf r} > 0$. We also define $[\bX\,|\,\by] \in \mathbb{R}^{n \times (d+1)}$ as the concatenation of the matrices $\bX$ and $\by$ and $\sigma_{\min}([\bX\,|\,\by])$ as it's minimum singular value.

\begin{definition}[Label-DP]\label{def:label-DP}
A randomized algorithm $\mc{A}$ taking a dataset as an input is $(\eps, \delta)$-label-DP for any $\eps, \delta \geq 0$ if for any two datasets $\bD$ and $\ol{\bD}$ which differ on the label of only a single example, and for any subset $S$ of outputs of $\mc{A}$,
$\Pr[\mc{A}(\bD) \in S] \leq e^\eps \Pr[\mc{A}(\bD') \in S] + \delta.$
\end{definition}
Our first aggregation mechanism \LFA given in Figure  \ref{alg:bag_aggregation} outputs $m$ examples, each by by randomly sampling $n$ i.i.d $N(0, 1)$ weights and taking a weighted sum of the feature-vectors as well as the labels. We prove (in Section \ref{sec:privacy_LFA} and Appendix \ref{appendix:lfa_utility}) the following result for \LFA.

\begin{algorithm}[H]
  \caption{\LFA}
  \label{alg:bag_aggregation}
  \begin{small}
\begin{algorithmic}
  \STATE {\bfseries Input:} $\{\bx_i, y_i\}_{i=1}^n$ and $m \in \mathbb{Z}^+$.
  \FOR{$j=1$ {\bfseries to} $m$}
  \STATE Sample iid $w_{ij} \sim N(0,1)$, $i = 1, \dots, n$.
  \STATE $\bx^{\sf AGG}_j := \sum_{i = 1}^n w_{ij}\bx_i$ ; $y^{\sf AGG}_j = \sum_{i = 1}^n w_{ij}y_i$.
  \ENDFOR
  \STATE {\bfseries Output:} $\mb{D}_{\tn{AGG}} := \left\{\bx^{\sf AGG}_j, y^{\sf AGG}_j\right\}_{j=1}^m$.
\end{algorithmic}
\end{small}
\end{algorithm}

\begin{theorem}\label{thm:LFA-main}
The output $\bD_{\sf AGG} = \left\{\bx^{\sf AGG}_j, y^{\sf AGG}_j\right\}_{j=1}^m$ of Algorithm \ref{alg:bag_aggregation} on a dataset $\bD = \{\bx_i, y_i\}_{i=1}^n$ such that $|y_i| \leq B_{\sf Y}$ for all $i \in [n]$ and $\opt(\bD, \mc{F}) \geq \gamma n$ is $(\eps, \delta)$-label-DP if
\[
n \geq \frac{16B_{\sf Y}^2}{\gamma}\max\left\{1, \frac{25\log^2\left(2/\delta\right)}{c_0^2\eps^2}\right\}\,, \quad \quad m \leq \frac{\eps\gamma n}{160B_{\sf Y}^2}\min\left\{1, \frac{2c_0\eps}{5 \log\left(2/\delta\right)}\right\}\,,
\]

where $c_0$ is the constant in Theorem \ref{thm:hanson-wright}. Furthermore, if $\br^* = {\sf argmin}_{\br \in \mc{F}_{B_{\sf r}}} \val(\bD_{\sf AGG}, \br^{\sf T}\bx)$ then $\val(\bD, \br^{*\sf T}\bx) \leq (1 + \theta)\opt(\bD, \mc{F}_{B_{\sf r}})$ for $\theta \in (0, 2)$ with probability at least
\begin{align}
    1 - \exp\bigg{(} & -\min\bigg{\{} \Omega(m^2)-O(\log n),  \Omega\left(m\theta^2\right) - O\left(d\log\left(\tfrac{m^2n(B_{\sf Y} + B_{\sf X}B_{\sf r})^2}{\theta\gamma}\right)\right)\bigg{\}}\bigg{)} \nonumber
\end{align}
\end{theorem}

Our next algorithm \LBA given in Figure \ref{fig-alg-Agg-Lin-LBA} randomly samples $m$ disjoint $k$-sized subsets of $[n]$, and iid $N(0,1)$ weights for each index in these subsets. Using these weights, for each subset it outputs the weighted sum of the feature vectors, along with the weighted sum of their labels, corresponding to the indices in that subset. We prove (in Section \ref{sec:privacynet} and Appendix \ref{sec:LBA-utility}) the following result for \LBA.

\begin{algorithm}[H]
  \caption{\LBA}
  \label{fig-alg-Agg-Lin-LBA}
  \scriptsize
\begin{algorithmic}
  \STATE {\bfseries Input:} $\mb{D} = \{(\bx_i, y_i)\}_{i=1}^n$ and $k, m \in \mathbb{Z}^+$.
  \STATE Sample $m$ disjoint $S_1, \dots, S_m \subseteq [n]$ s.t. $|S_j| = k$. 
  \STATE Construct $\bX_j = [\bx_i^{\sf T}]_{i \in S_j}$ and $\by_j = [y_i^{\sf T}]_{i \in S_j}$ for all $j \in [m]$.
  \FOR{$j=1$ {\bfseries to} $m$}
  \STATE Sample $\bw_j \sim N(0,1)^k$. 
  \STATE Compute $\bx^{\sf AGG}_j := \bw^{\sf T}\bX_j$ ; $y^{\sf AGG}_j = \bw_j^{\sf T}\by_j$.
  \ENDFOR
  \STATE {\bfseries Output:} $\bD_{\sf AGG} = \{(\bx^{\sf AGG}_j, y^{\sf AGG}_j)\}_{j=1}^m$ and $S_1, \dots, S_m$.
\end{algorithmic}
\end{algorithm}

\begin{theorem}\label{thm-intro-main1}
The output $\bD_{\sf AGG} = \{(\bx^{\sf AGG}_j, y^{\sf AGG}_j)\}_{j=1}^m$ and $S_1, \dots, S_m$ of Algorithm \ref{fig-alg-Agg-Lin-LBA} on a dataset $\bD = (\bX, \by) \in \R^{n \times d} \times \in \R^{n \times 1}$ such that $|\by_i| \leq B_{\sf Y}$ and $\|\bx_i\|_2^2 \leq B_{\sf X}^2$ for all $i \in [n]$, and $\sigma_{\sf min}([\bX\,|\,\by])^2 \geq n\lambda^*$ is $(\eps, \delta)$-label-DP for $\eps \leq 10$ if
\begin{align}
     k \geq \max\bigg{\{} & 2, 1 + \left(\tfrac{(n-1)(B_{\sf Y}^2 + B_{\sf X}^2)}{n\lambda^* - B_{\sf Y}^2 + B_{\sf X}^2}\right)\left(\tfrac{\log((2d+2)/\delta)}{1 - \log(2)}\right),  \tfrac{640B_{\sf Y}^2(n-1)}{\eps(n\lambda^* - B_{\sf Y}^2 - B_{\sf X}^2)}, \tfrac{1600B_{\sf Y}^2\log^2(4/\delta)(n-1)}{c_0^2\eps^2(n\lambda^* - B_{\sf Y}^2 - B_{\sf X}^2)}\bigg{\}} \nonumber
\end{align}
where $c_0$ is the constant in Theorem \ref{thm:hanson-wright}. Furthermore, if $\br^* = {\sf argmin}_{\br \in \mc{F}_{B_{\sf r}}}\val(\bD_{\sf AGG}, \br^{\sf T}\bx)$ then $\val(\bD, \br^{*\sf T}\bx) \leq (1 + \theta)\opt(\bD, \mc{F}_{B_{\sf r}})$ for $\theta \in (0, \tfrac{1}{2})$ with probability at least
 \begin{align}
      1 - & \exp\bigg{(} -\min\bigg{\{}\Omega(m^2) - O(\log k) ,  \Omega\left(\tfrac{mk\theta^2\gamma^2}{(B_{\sf Y} + B_{\sf X}B_{\sf r})^4}\right)- O\left(d\log\left(\tfrac{m^2k(B_{\sf Y} + B_{\sf X}B_{\sf r})^2}{\theta\gamma}\right)\right)\bigg{\}}\bigg{)} \nonumber    
 \end{align}
 \end{theorem}

\subsection{Discussion of Results} \label{sec:discussion}
As mentioned in Section \ref{sec:related-work},~\citep{Sheffet-old} also studies the \LFA mechanism (see Section 3, Theorem 2 of \cite{Sheffet-old}), viewing it as a Johnson-Lindenstrauss projection via a random matrix $\mb{R} \in N(0,1)^{m\times n}$ of the $n\times (d+1)$ matrix whose $i$th row is the $i$th feature-vector concatenated with its label. The output is identical to that of \LFA and its DP guarantee is w.r.t to the perturbation of an entire (feature-vector, label) pair. Their work shows that for a constant $\eps$, $m$ can be taken up to $O(w^4/ B^2)/(\log(1/\delta))$ and $B$ is the upper bound on $\ell_2$ norm of any feature-vector concatenated with its label and $w$ is the minimum singular-value of the matrix of feature-vectors concatenated with their labels. In comparison, we obtain $O(\gamma n/ B_{\sf Y}^2)/(\log(1/\delta)$ where $\gamma$ is the lower bound on the linear regression loss and $B_{\sf Y}$ is a bound on the magnitude of the labels. We observe that $\gamma \geq w^2/n$ using the following argument.
\begin{align}
    \gamma & = \tfrac{1}{n}\min_{\br \in \mathbb{R}^d}\|\by - \bX\br\|_2^2 \nonumber \\
    & = \tfrac{1}{n}\min_{\br \in \mathbb{R}^d}\sqrt{1 + \|\br\|_2^2}\left\|\tfrac{\by}{\sqrt{1 + \|\br\|_2^2}} - \tfrac{\bX\br}{\sqrt{1 + \|\br\|_2^2}}\right\|_2^2 \nonumber \\
    & \geq \tfrac{1}{n}\min_{\br \in \mathbb{R}^d}\left\|\tfrac{\by}{\sqrt{1 + \|\br\|_2^2}} - \tfrac{\bX\br}{\sqrt{1 + \|\br\|_2^2}}\right\|_2^2 = \tfrac{1}{n}w^2 \nonumber
\end{align}

Since \( B \geq B_{\sf Y} \), \((w^2/nB)^2\) is smaller than \(\gamma/B_{\sf Y}\). The result in \citep{Sheffet-old} allows \( m \) to be at most \( O((w^2/nB)^2 n^2 / \log(1/\delta)) \), whereas our result allows \( m \) to be at most \( O((\gamma/B_{\sf Y}) n / \log(1/\delta)) \). Thus, while the result in \citep{Sheffet-old} permits \( m \) to grow as \( O(n^2) \), our result allows \( m \) to grow as \( O(n) \). However, since the value of \((w^2/nB)^2\) is much smaller than \(\gamma/B_{\sf Y}\) in practice, our result performs significantly better. We empirically demonstrate this in Section \ref{sec:experiments} on real datasets, showing that the \LFA upper bound on \( m \) is substantially better. The minimum eigenvalue assumption is used by \cite{Sheffet-old} to prove the preservation of ridge regression utility. Since we do not make that assumption, \LFA admits utility preservation for bounded norm linear regressors.

The \LBA mechanism does however assume a minimum singular value bound, but has the added benefit that the $\mb{R}$ matrix in this case is sparse -- the non-zero entries in the $i$th row is the support of the $i$th $k$-sized bag. Therefore, $\mb{R}$ has only $mk$ non-zero entries, unlike the dense $\mb{R}$ of \LFA and in Theorem 2 of \cite{Sheffet-old}. The minimum eigenvalue ensures that the regression loss within a bag is approximately the same as the global regression loss, with high probability. Since the bags are disjoint,  any perturbation of a single label affects at most one of the $m$ aggregates and and thus the privacy analysis is similar to that of \LFA for $k$ feature-vectors and $m=1$.

\section{Overview of our Techniques}\label{sec:overview}

{\bf Theorem \ref{thm:LFA-main}.} Let us for simplicity consider just one bag from Algorithm \ref{alg:bag_aggregation} consisting of all the feature-vectors and labels of the dataset. Let $\bX$ be the matrix whose rows are the feature vectors of the dataset and $\by$ be the corresponding vector of their labels. Let $\bw \sim N(0,1)^n$ be the weight vector sampled yielding $(\bx^{\sf AGG})^{\sf T} := \bw^{\sf T}\bX$ and $\by^{\sf AGG} := \bw^{\sf T}\by$ as the output for this bag. Suppose for the moment that $\by$ belongs to the column-space of $\bX$, i.e. there is $\mb{h}$ s.t. $\by = \bX\mb{h}$, and thus $\by^{\sf AGG} = \bw^{\sf T}\by = \bw^{\sf T}\bX\mb{h} = (\bx^{\sf AGG})^{\sf T}\mb{h}$. In this case therefore, given  $(\bx^{\sf AGG})^{\sf T} := \bw^{\sf T}\bX$ there is no randomness in the output $\by^{\sf AGG}$ and any change from $\by$ to $\ol{\by}$ in the $i^{\sf th}$ coordinate can be detected as long as $w_i \neq 0$ (see Appendix \ref{sec:yincolspaceX} for an explanation). 

In our case however, $\by$ is sufficiently far from being in the column-space of $\bX$ which is ensured by the lower bound on the linear regression loss. Our key finding is that for any choice of $(\bx^{\sf AGG})^{\sf T}$, the conditional distribution on $\bw^{\sf T}\by$ is a $\mu$-mean Gaussian with variance precisely equalling $\opt(\bD, \mc{F})$ (see Eqn. \eqref{eq:sigmabd_aggregate_main}). When $\by$ changes by a single label to $\by'$, while the distribution of $\bw^{\sf T}\bX$ does not change, the distribution of $\bw^{\sf T}\by'$ (conditioned on $\bw^{\sf T}\bX$) has a slightly different mean and variance. 

Notice that since we are restriction ourselves to label-DP, the guarantee must be independent of the input dimension $d$. The proof in \citep{Sheffet-old} proceeds by representing the difference between matrices of neighbouring datasets as $\mb{e}_i^{\sf T}(\bv - \ol{\bv})$. This formulation enables the use of the Sherman-Morrison Lemma and the Matrix-Determinant Lemma to obtain tight bounds. In contrast, for proving a label-DP guarantee, the difference between neighboring datasets is represented as $\mb{E}_{i(d+1)}(v-\ol{v})$, where $\mb{E}_{i(d+1)}$ is the basis matrix and $v$ and $\ol{v}$ are scalars. Applying the Sherman-Morrison Lemma and Matrix-Determinant Lemma to this formulation does not yield tight bounds. This limitation prevents both: (i) the use of the lower bound on the minimum loss instead of that on the minimum singular value of the data matrix, and (ii) the getting a guarantee independent of the input dimension $d$.

The primary novelty of our proof lies in reducing a $(d+1)$-dimensional random variable $(\bw^{\sf T}\bX, \bw^{\sf T}\by)$ to a $2$-dimensional random variable (see equation \eqref{eqn:pdf_ratio}). This reduction enables us to obtain a much tighter bound that is independent of the dimension $d$. Additionally, it allows the use of lower bound on the minimum loss.

For the theoretical utility bound we first show using Gaussian concentration bounds that any \emph{bad} linear regressor w.h.p. remains bad for the \LFA output. Next, we use a union-bound over a fine-grained net of the regressors in  $\mc{F}_{B_{\sf r}}$ (see \citep{Vershynin-book})  to obtain a high probability bound for all bad regressors in $\mc{F}_{B_{\sf r}}$. Using the lower bound on the optimum, we obtain a $(1 \pm o(1))$ multiplicative bound on the \LFA loss w.r.t. the loss on the input dataset for all linear-regressors in $\mc{F}_{B_{\sf r}}$, except with probability $\approx \exp(-\Omega(m))$.

{\bf Theorem \ref{thm-intro-main1}.}  Since the $m$ bags, each $k$-sized, are disjoint, a perturbation in one label only affects at most one of the aggregates. Thus, the privacy analysis boils down to that for \LFA with $m = 1$. However, we need to show a high probability loss bound within the sampled bags, and for this we use the minimum eigenvalue bound along with an application of Matrix Chernoff inequality. 

The utility proof uses the Hoeffding as well as Gaussian concentration bounds along with a union bound over net argument, and is similar to that for \LFA. Section \ref{sec:privacynet} and Appendix \ref{sec:LBA-utility} formally prove the privacy and utility guarantees for \LBA respectively.

{\bf Experiments.} We empirically demonstrate the utility preservation of linear regression on \LFA with realistic privacy budgets on the large-scale 1940 US Census dataset~\citep{IPUMS-USA} and the Criteo Sponsored Search dataset~\cite{tallis2018reacting}. The results include the privacy vs. empirical utility trade-off and are presented in Section \ref{sec:experiments}. We also evaluate \LBA and \LLP for linear and neural regression respectively on two large-scale regression datasets. Our experiments reinforce our theoretical guarantees for \LBA and \LLP, demonstrating that they preserve the utility even with a small number of sampled bags of various sizes (see Appendix \ref{sec:additional_exp}). 

We wish to emphasize that the primary goal of this work is to quantify the privacy achieved through aggregation alone. Unlike other label-differential privacy (label-DP) methods e.g. \citep{ghazi2022regression} and \citep{esfandiari2022label} which rely on the addition of noise or probabilistic resampling, our approach focuses solely on aggregation with no or little additive noise or perturbations. Consequently, we omit experimentally comparing our work with previous methods, as it is not directly relevant to our motivation.

\section{Preliminaries} \label{sec:prelims-main}

\begin{theorem}[Hanson-Wright inequality, Theorem 1.1 in \citep{rudelson2013hanson}]\label{thm:hanson-wright}
Let $\bX = (X_1, \dots, X_n) \in \R^n$ be a random vector of independent components $X_i$ which satisfy $\E X_i = 0$ and $\|X_i\|_{\psi_2} \leq K$. Let $\bA$ be an $n \times n$ matrix. Then for every $t \geq 0$,
\begin{align}
    \Pr[|\bX^{\sf T}\bA\bX - & \E \bX^{\sf T}\bA\bX| > t] \leq \nonumber 2\exp\left(-c_0\min\left\{\tfrac{t^2}{K^4\|\bA\|_F^2}, \tfrac{t}{K^2\|\bA\|_2}\right\}\right)\nonumber
\end{align}
where $c_0 > 0.145$ is an absolute constant (see \citep{moshksar2021absolute} for the lower bound on $c_0$)
\end{theorem}

We prove the following variation of the Matrix Chernoff bound for sampling rows of a matrix uniformly at random without replacement in Appendix \ref{appendix:matrix_chernoff_wo_replacement}. 
\begin{theorem}\label{lem:matrix_chernoff_wo_rreplacement}
Consider a finite sequence $\{\bX_i\}_{i=1}^k$ of Hermitian matrices of dimension $d$ sampled uniformly at random without replacement from a fixed set. Let \begin{align}
    \bX_i \geq 0 \qquad \lambda_{\max}(\bX_i) \leq R \nonumber
\end{align}
Define $\mu_{\min} := \lambda_{\min}\left(\sum_{i = 1}^k\E \bX_i\right)$. Then,
\begin{align}
    \Pr\left[\lambda_{\min}\left(\sum_{i=1}^k\bX_i\right) \leq (1 - \delta)\mu_{\min}\right] \leq d \left[\tfrac{e^{-\delta}}{(1 - \delta)^{1 - \delta}}\right]^{\tfrac{\mu_{\min}}{R}}\nonumber
\end{align}
\end{theorem}

\section{Privacy analysis of \LFA} \label{sec:privacy_LFA}
In this section prove the privacy result in Theorem \ref{thm:LFA-main}. We defer the proof of the utility result to Appendix \ref{appendix:lfa_utility}.

Define matrices $\bX := [\bx_i^{\sf T}]_{i=1}^n \in \mathbb{R}^{n \times d}$, $\by := [y_i]_{i=1}^n \in \mathbb{R}^{n \times 1}$ and $\bW := {[w_{ij}]_{i=1}^n}_{j=1}^m \in \mathbb{R}^{n \times m}$. Fix $\bX$ for the rest of the proof. 

Let $\mc{D}$ be the distribution of $\bW^{\sf T}\bX$, $\mc{D}_{\bZ, \by}$ be the distribution of $\bW^{\sf T}\by$ conditioned on $\bW^{\sf T}\bX$ taking the value $\bZ$ and $\mc{D}_{\by}$ be the distribution of $(\bW^{\sf T}\bX, \bW^{\sf T}\by)$.

Let $\bX = \bU\bm{\Sigma}\bV^{\sf T}$ be the singular value decomposition where $\bU \in \R^{n \times n}$ and $\bV \in \R^{d \times d}$ are unitary matrices and $\bm{\Sigma} \in \R^{n \times d}$ is a diagonal matrix with non zero singular values $\sigma_{1} \geq \dots \geq \sigma_{\ol{d}} \geq 0$ where $\ol{d} \leq d$. Define
$$
\tilde{\by} = \bU^{\sf T}\by \qquad \tilde{\bW} = \bU^{\sf T}\bW \qquad \tilde{\bZ} = \bZ\bV
$$
Notice that $\mc{D}_{\bZ, \by}$ is the distribution of $\tilde{\bW}^{\sf T}\tilde{\by}$ conditioned on $\tilde{\bW}^{\sf T}\bm{\Sigma}$ taking the value $\tilde{\bZ}$. Since $\bU$ is unitary, $\tilde{\bW} \sim N(0, 1)^{n \times m}$. Thus, $\mc{D}_{\bZ, \by}$ is a gaussian distribution with the mean $\bm{\mu} \in \R^{m}$ and variance of $\sigma^2\bm{I}_{m \times m}$ where
$$
    \mu_j = \sum_{i=1}^{\ol{d}}\left(\tilde{Z}_{ji}/\sigma_i\right)\tilde{y}_i \quad \forall\, j \in [m] \,\,\text{   and   }\,\,\sigma^2 = \sum_{i=\ol{d}+1}^{n}\tilde{y}_i^2 \nonumber
$$
Observe that since exactly the first $d' \leq d$ singular values are non-zero, $\bm{\Sigma}\tilde{\bv}$ can be non-zero only in the first $d'$ coordinates for any $\tilde{\bv} \in \R^d$. Thus, the minima over $\tilde{\bv}$ of $\|\tilde{\by} - \bm{\Sigma}\tilde{\bv}\|_2$ is obtained exactly when $\bm{\Sigma}\tilde{\bv}$ cancels out $\tilde{y}$ in the first $d'$ coordinates. Thus, substituting $\tilde{\bv} = \bV^{\sf T}\bv$ we obtain
\begin{align}
    \sigma^2 = \sum_{i=d'+1}^{n}\tilde{y}_i^2 & = \underset{\tilde{\bv} \in \R^d}{\inf}\|\bU(\tilde{\by} - \bm{\Sigma}\tilde{\bv})\|_2^2  = \underset{\bv \in \R^d}{\inf}\|\by - \bX\bv\|_2^2 \geq \gamma n \label{eq:sigmabd_aggregate_main}
\end{align}
Let $\ol{\by} \in \R^{n \times 1}$ such that $\|\ol{\by} - \by\|_0 = 1$. Let $\tilde{\ol{\by}} = \bU^{\sf T}\ol{\by}$ and $\mc{D}_{\bZ, \ol{\by}} \sim N(\ol{\mu}, \ol{\sigma})$. 
We have the following inequality.
\begin{align}
    \ol{\sigma} & = \underset{\bw \in \R^d}{\inf}\|\ol{\by} - \bX\bw\|_2 \leq \underset{\bw \in \R^d}{\inf}\|\by - \bX\bw\|_2 + \|\by - \ol{\by}\|_2 \leq \sigma + 2B_{\sf Y} \label{eq:olsigmabd_plus_1} \\
    \ol{\sigma} & = \underset{\bw \in \R^d}{\inf}\|\ol{\by} - \bX\bw\|_2  \geq \underset{\bw \in \R^d}{\inf}\|\by - \bX\bw\|_2 - \|\ol{\by} - \by\|_2 \geq \sigma - 2B_{\sf Y} \label{eq:olsigmabd_minus_1}
\end{align}

The logarithm of the ratio of pdfs at a point $(\bZ, \mb{t}) \in \R^{m \times (d+1)}$ is
\begin{align}
    & = \log\left(\frac{\underset{\mc{D}_{\by}}{\Pr}[(\bZ, \mb{t})]}{\underset{\mc{D}_{\ol{\by}}}{\Pr}[(\bZ, \mb{t})]}\right) = \log\left(\frac{\underset{\mc{D}_{\bZ, \by}}{\Pr}[\mb{t}]}{\underset{\mc{D}_{\bZ, \ol{\by}}}{\Pr}[\mb{t}]}\right) \nonumber \\
    & = \log\left(\prod_{j=1}^{m}\left(\frac{\tfrac{1}{\sqrt{2\pi}\sigma}\exp\left(-\tfrac{1}{2}\left(\tfrac{t_j - \mu_j}{\sigma}\right)^2\right)}{\tfrac{1}{\sqrt{2\pi}\ol{\sigma}}\exp\left(-\tfrac{1}{2}\left(\tfrac{t_j - \ol{\mu}_j}{\ol{\sigma}}\right)^2\right)}\right)\right) \nonumber \\
    & \leq \tfrac{1}{2}\left[\sum_{j=1}^m\left(\left(\tfrac{t_j - \ol{\mu}_j}{\ol{\sigma}}\right)^2 - \left(\tfrac{t_j - \mu_j}{\sigma}\right)^2\right) + m\tfrac{\ol{\sigma}^2}{\sigma^2} - m\right] \label{eqn:pdf_ratio}
\end{align}

Notice that \( t_j = \sum_{i=1}^{n} \tilde{w}_{ij} \tilde{y}_i \), \( \mu_j = \sum_{i=1}^{\ol{d}} \tilde{w}_{ij} \tilde{y}_i \), and \( \ol{\mu}_j = \sum_{i=1}^{\ol{d}} \tilde{w}_{ij} \tilde{\ol{y}}_i \). When \( (\bZ, \mb{t}) \sim \mc{D}_{\by} \) for all \( j \in [m] \), the random variable \( R_j := (t_j - \mu_j)/\sigma \) follows a standard normal distribution \( N(0, 1) \). Additionally,

\begin{align}
\tfrac{t_j - \ol{\mu}_j}{\ol{\sigma}} = \tfrac{t_j - \mu_j}{\ol{\sigma}} + \tfrac{\mu_j - \ol{\mu}_j}{\ol{\sigma}} \nonumber
\end{align}

where \( \Var[\mu_j - \ol{\mu}_j] = \sum_{i=1}^{\ol{d}} (\tilde{y}_i - \tilde{\ol{y}}_i)^2 \) and \( \text{Cov}[t_j - \mu_j, \mu_j - \ol{\mu}_j] = 0 \). This gives:

\begin{align}
\tfrac{t_j - \ol{\mu}_j}{\ol{\sigma}} = \left(\tfrac{\sigma}{\ol{\sigma}}\right) R_j + \left(\tfrac{\sqrt{\sum_{i=1}^{\ol{d}} (\tilde{y}_i - \tilde{\ol{y}}_i)^2}}{\ol{\sigma}}\right) S_j \nonumber
\end{align}

where \( S_j \sim N(0, 1) \) and \( S_j \) is independent of \( R_j \). Furthermore, for any \( j \neq j' \), we have \( R_j \perp R_{j'} \), \( S_j \perp S_{j'} \), and \( R_j \perp S_{j'} \) since \( \tilde{\bW} \sim N(0, 1)^{n \times m} \).

Define the random vector \( \mb{T} := [S_1, R_1, \dots, S_m, R_m]^{\sf T} \). Let \( \sigma_1 = \sigma/\ol{\sigma} \) and \( \sigma_2 = \left(\sqrt{\sum_{i=1}^{\ol{d}} (\tilde{y}_i - \tilde{\ol{y}}_i)^2}/\ol{\sigma}\right) \). Define the block diagonal matrix \( \bA \) with each block:

\begin{align}
\bA_0 = \begin{bmatrix}
    \sigma_1^2 - 1 & \sigma_1 \sigma_2 \\
    \sigma_1 \sigma_2 & \sigma_2^2
\end{bmatrix} \nonumber
\end{align}

of size \( 2 \times 2 \). Thus, we have:

\begin{align}
\sum_{j=1}^m \left( \left( \tfrac{t_j - \ol{\mu}_j}{\ol{\sigma}} \right)^2 - \left( \tfrac{t_j - \mu_j}{\sigma} \right)^2 \right) = \mb{T}^{\sf T} \bA \mb{T} \nonumber
\end{align}

Taking the expectation gives $
\E[\mb{T}^{\sf T} \bA \mb{T}] = m(\sigma_1^2 + \sigma_2^2 - 1)
$. Thus, the expression in \eqref{eqn:pdf_ratio} simplifies to:

\begin{align}
= \frac{1}{2} \left( \mb{T}^{\sf T} \bA \mb{T} - \E[\mb{T}^{\sf T} \bA \mb{T}] + m \left( (\sigma_1 - \tfrac{1}{\sigma_1})^2 + \sigma_2^2 \right) \right) \nonumber
\end{align}

Notice that \( \sigma_2^2 \leq \frac{16 B_{\sf Y}^2}{\gamma n} \) because \( \sum_{i=1}^{\ol{d}} (\tilde{y}_i - \tilde{\ol{y}}_i)^2 \leq 4B_{\sf Y}^2 \) and \( \ol{\sigma}^2 \geq \frac{\gamma n}{4} \) for \( n \geq \frac{16 B_{\sf Y}^2}{\gamma} \). Moreover, \( (\sigma_1 - \frac{1}{\sigma_1})^2 \leq \frac{64 B_{\sf Y}^2}{\gamma n} \) for \( n \geq \frac{16 B_{\sf Y}^2}{\gamma} \). Therefore, we have:

\begin{align}
m \left( (\sigma_1 - \tfrac{1}{\sigma_1})^2 + \sigma_2^2 \right) \leq m \left( \tfrac{80 B_{\sf Y}^2}{\gamma n} \right) \leq \tfrac{\eps}{2} \nonumber
\end{align}

whenever \( m \leq \frac{\eps \gamma n}{160 B_{\sf Y}^2} \) and \( n \geq \frac{16 B_{\sf Y}^2}{\gamma} \).

We apply Theorem \ref{thm:hanson-wright} to establish a high-probability upper bound of \( \eps/2 \) on \( \frac{1}{2} \left( \mb{T}^{\sf T} \bA \mb{T} - \E[\mb{T}^{\sf T} \bA \mb{T}] \right) \). Given that \( \|S_j\|_{\Psi_2} = \|R_j\|_{\Psi_2} = 1 \), \( \|\bA\|_F^2 = r \|\bA_0\|_F^2 \), and \( \|\bA\|_2 = \|\bA_0\|_2 \), we find:
\begin{align}
    \Pr[|\mb{T}^{\sf T}\bA\mb{T} - & \E[\mb{T}^{\sf T}\bA\mb{T}]| \geq \eps ] \leq 2\exp\left(-c_0\min\left\{\tfrac{\eps^2}{m\|\bA_0\|_F^2}, \tfrac{\eps}{\|\bA_0\|_2}\right\}\right) \label{eqn:hanson_wright_on_full_bags}
\end{align}
Next, we derive an upper bound for \( \|\bA_0\|_F^2 = (\sigma_1^2 - 1)^2 + 2\sigma_1^2\sigma_2^2 + \sigma_2^4 \). With \( n \geq \frac{16 B_{\sf Y}^2}{\gamma} \), we have \( \sigma \geq 4B_{\sf Y} \) which implies \( \ol{\sigma} \geq \frac{\sigma}{2} \). Thus we get (i)  \( (\sigma_1^2 - 1)^2 \leq 256 \frac{B_{\sf Y}^2}{\gamma n}\); (ii) \( 2\sigma_1^2 \sigma_2^2 \leq 128 \frac{B_{\sf Y}^2}{\gamma n}\); and (iii)  \( \sigma_2^4 \leq 16 \frac{B_{\sf Y}^2}{\gamma n}\). Combining these, we find \( \|\bA_0\|_F^2 \leq 400 \frac{B_{\sf Y}^2}{\gamma n} \).

Next, we bound the failure probability terms in equation \eqref{eqn:hanson_wright_on_full_bags}:

1. From \( 2 \exp \left( -\frac{c_0 \eps}{\|\bA_0\|_2} \right) \leq \delta \,:\, n \geq \frac{400 B_{\sf Y}^2 \log^2(2/\delta)}{c_0^2 \gamma \eps^2}\):

2. From \( 2 \exp \left( -\frac{c_0 \eps^2}{m \|\bA_0\|_F^2} \right) \leq \delta \,:\, m \leq \frac{c_0 \eps^2 \gamma n}{400 B_{\sf Y}^2 \log(2/\delta)}\):

This concludes the proof.

\section{Privacy Analysis for \LBA}\label{sec:privacynet}
In this section we prove the privacy result in Theorem \ref{thm-intro-main1}. The utility analysis is deferred to Appendix \ref{sec:LBA-utility}.

For the rest of the proof, define $\ba_i \in \mathbb{R}^{d+1}$ as the concatenation of vector $\bx_i$ with its label $y_i$; $\bA_j \in \mathbb{R}^{k \times (d+1)}$ as the concatenation of matrices $\bX_j$ and $\by_j$, and $\bA \in \mathbb{R}^{n \times (d+1)}$ as the concatenation of matrices $\bX$ and $\by$. Let \( \ol{\by} \in \mathbb{R}^{n \times 1} \) such that \( \|\by - \ol{\by}\|_0 = 1 \). Without loss of generality, assume \( i_0 \) is the index where \( \ol{\by} \) differs from \( \by \). If we condition on the event that \( i_0 \in S_{j_0} \), the distribution of \( (\bw_j^{\sf T} \bX_j, \bw_j^{\sf T} \by_j) \) remains unchanged for all \( j \neq j_0 \) since the bags are non-overlapping. Additionally, we can bound the distribution of \( (\bw_{j_0}^{\sf T} \bX_{j_0}, \bw_{j_0}^{\sf T} \by_{j_0}) \). Using the matrix Chernoff bound, we obtain a high-probability lower bound on the minimum singular value of \( \bA \). Specifically, we have:
\begin{align}
\lambda_{\sf min}(\bA_{j_0}^{\sf T}\bA_{j_0}) \geq \lambda_{\sf min}\left(\sum_{i \in S_{j_0} - \{i_0\}}\ba_i\ba_i^{\sf T}\right). \nonumber
\end{align}

Further, by Weyl's inequality, we have:
\begin{align}
\underset{{S_{j_0} - \{i_0\}}}{\E}\left[\sum_{i \in S_{j_0} - \{i_0\}} \ba_i \ba_i^{\sf T}\right] &  = \tfrac{k-1}{n-1} \lambda_{\min}(\bA^{\sf T} \bA - \ba_i \ba_i^{\sf T})  = (k-1) \tfrac{n \lambda^* - B^2}{n-1} \nonumber
\end{align}
where \( B^2 = B_{\sf Y}^2 + B_{\sf X}^2 \). Applying matrix Chernoff bound:
\begin{align}
\Pr\left[\lambda_{\min}\left(\sum_{i \in S_{j_0} - \{i_0\}} \ba_i \ba_i^{\sf T}\right) \leq \tfrac{(k-1)(n \lambda^* - B^2)}{2(n-1)}\right]
\leq (d+1)\left(2/e\right)^{\frac{(k-1)(n \lambda^* - B^2)}{(n-1) B^2}} \nonumber
\end{align}
Thus, the failure probability is at most \( \delta/2 \) if 
\begin{align}
k \geq 1 + \tfrac{(n-1)(B_{\sf Y}^2 + B_{\sf X}^2)}{n \lambda^* - B_{\sf Y}^2 + B_{\sf X}^2}\cdot\tfrac{\log((2d+2)/\delta)}{1 - \log(2)} \nonumber
\end{align}
Let \( \mc{D}_{\by} \) and \( \mc{D}_{\ol{\by}} \) be the output distributions for \( (\bX, \by) \) and \( (\bX, \ol{\by}) \). The logarithm of the ratio of PDFs at \( (\bZ, \mb{t}) \in \R^{m \times (d+1)} \) for the bags \( \mb{P} = \{P_1, \ldots, P_m\} \) (where \( i_0 \in P_{j_0} \)) is:
\begin{align}
    \log\left(\tfrac{\Pr_{\mc{D}_{\by}}[(\bZ, \by, \mb{P})]}{\Pr_{\mc{D}_{\ol{\by}}}[(\bZ, \by, \mb{P})]}\right) & = \log\left(\tfrac{\Pr_{\mc{D}_{\by}}[\mb{P}]\prod_{j=1}^m\Pr_{\mc{D}_{\by}}[(\bz_j,t_j)\,|\,P_j]}{\Pr_{\mc{D}_{\ol{\by}}}[\mb{P}]\prod_{j=1}^m\Pr_{\mc{D}_{\ol{\by}}}[(\bz_j,t_j)\,|\,P_j]}\right)  = \log\left(\tfrac{\Pr_{\mc{D}_{\by}}[(\bz_{j_0}, t_{j_0})\,|\,P_{j_0}]}{\Pr_{\mc{D}_{\ol{\by}}}[(\bz_{j_0}, t_{j_0})\,|\,P_{j_0}]}\right) \nonumber
\end{align}
Assuming the lower bound on the minimum singular value of \( \bA_{j_0} \) holds, we use Theorem \ref{thm:LFA-main} with \( m = 1 \) and \( \gamma = \frac{1}{4}\left(\frac{n \lambda^* - B^2}{n-1}\right) \). This leads to an upper bound of \( \eps \) for the ratio with probability at least \( 1 - \frac{\delta}{2} \) if:
\begin{align}
k \geq \max\bigg{\{}&\tfrac{64 B_{\sf Y}^2 (n-1)}{n \lambda^* - B_{\sf Y}^2 - B_{\sf X}^2}, \tfrac{640 B_{\sf Y}^2 (n-1)}{\eps(n \lambda^* - B_{\sf Y}^2 - B_{\sf X}^2)},  \tfrac{1600 B_{\sf Y}^2 \log^2(4/\delta)(n-1)}{c_0^2 \eps^2(n \lambda^* - B_{\sf Y}^2 - B_{\sf X}^2)}\bigg{\}} \nonumber
\end{align}
This concludes the proof.

\section{Experimental results}\label{sec:experiments}
We experimentally evaluate our \LFA result against the results of \citep{Sheffet-old} on the 1940 US Census Data~\citep{IPUMS-USA} and the Criteo Sponsored Search Dataset~\citep{tallis2018reacting}. The US Census dataset contains $131,903,910$ datapoints, and the task is to predict the number of working hours of a person given access to features such as the gender, marital status, number of children, school attendance, employment status and type of industry they work in. The Criteo Dataset contains $1,732,721$ datapoints, and the task is to predict the sales amount of an advertisement using features such as delay in conversion, number of clicks within a week and various product categories. Further details of the dataset can be found in Appendix \ref{sec:additional_exp}. For each categorical feature, we perform a random projection on $\log k$ dimensional space if $k$ is its one-hot dimension.

The minimum squared singular value on the concatenation of $\bX$ and $\by$ on the US Census dataset is $7.57e-5$ and the same on the Criteo Dataset is $7.83$. The bound on the norm of concatenation of $\bx^{(i)}$ and $y^{(i)}$ is $121.73$ for the Census Dataset and $505.89$ for the Criteo Dataset. Since $\omega^2/B^2 \leq 1$ (refer to Theorem 11 in \citep{Sheffet-old}), even a single bag sampled cannot provide a differential privacy guarantee. On the other hand, our results allow us to offer label differential privacy guarantee using $\delta$ as $1e-8$ for the US Census Dataset and $1e-5$ for the Criteo Dataset, even after sampling sufficient number of bags to obtain good utility. The optimum of mse-linear regression on the US Census dataset is $\approx 64.22$ and the range of the target variable is $51 (B_{\sf Y} = 25.5)$. Similarly, The optimum of mse-linear regression on the Criteo dataset is $\approx 195.5$ and the range of the target variable is $254 (B_{\sf Y} = 127)$. 
In Tables \ref{tab:num_bags_census} and \ref{tab:num_bags_criteo}, we show the variation of the upper bounds on the number of bags with $\eps$ using $\delta = 1e-8$ and $\delta=1e-5$ for the US Census and the Criteo Sponsored Search Datasets respectively.

\begin{table}[ht]
\caption{Number of bags vs $\eps$ for both datasets}
\label{tab:num_bags}
\centering
\begin{subtable}{0.45\linewidth}
    \centering
    \caption{US Census}
    \label{tab:num_bags_census}
    \small
    \begin{tabular}{|c|c|}
    \toprule
    $\eps$ & Number of bags \\ \midrule
    1     & 247   \\ 
    2     & 988   \\ 
    3     & 2223  \\ 
    4     & 3853  \\ 
    5     & 6176  \\ 
    6     & 8894  \\ 
    7     & 12106 \\ 
    8     & 15812 \\ 
    9     & 20012 \\ 
    10    & 24706 \\ 
    11    & 29894 \\ 
    12    & 35577 \\ 
    13    & 41753 \\ 
    14    & 48424 \\ 
    \bottomrule
    \end{tabular}
\end{subtable}\hfill
\begin{subtable}{0.45\linewidth}
    \centering
    \caption{Criteo Sponsored Search}
    \label{tab:num_bags_criteo}
    \small
    \begin{tabular}{|c|c|}
    \toprule
    $\eps$ & Number of bags \\ \midrule
    6     & 89  \\ 
    7     & 122 \\ 
    8     & 159 \\ 
    9     & 202 \\ 
    10    & 249 \\ 
    11    & 302 \\ 
    12    & 359 \\ 
    13    & 422 \\ 
    14    & 489 \\ 
    15    & 562 \\ 
    \bottomrule
    \end{tabular}
\end{subtable}
\end{table}

We experimentally show linear regression utility preservation on the US Census and Criteo datasets with modest privacy budgets for \LFA. To illustrate the trend of utility with increasing privacy budget $\eps$, we sample an aggregated dataset containing as many aggregates as computed in Table \ref{tab:num_bags_census}, compute the optimizer of the aggregated dataset and its loss on the instance dataset with increasing privacy budget.

\begin{figure}[htb]
\centering

\begin{minipage}[t]{0.48\textwidth}
    \centering
    \includegraphics[width=\textwidth]{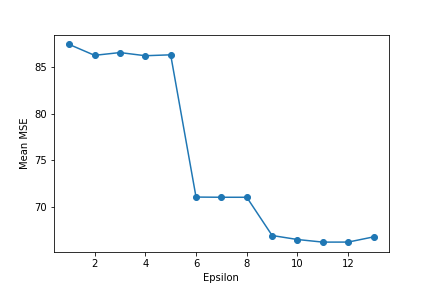}
    \caption{$\eps$ versus MSE for US Census Dataset}
    \label{fig:eps_opt}
\end{minipage}
\hfill
\begin{minipage}[t]{0.48\textwidth}
    \centering
    \includegraphics[width=\textwidth]{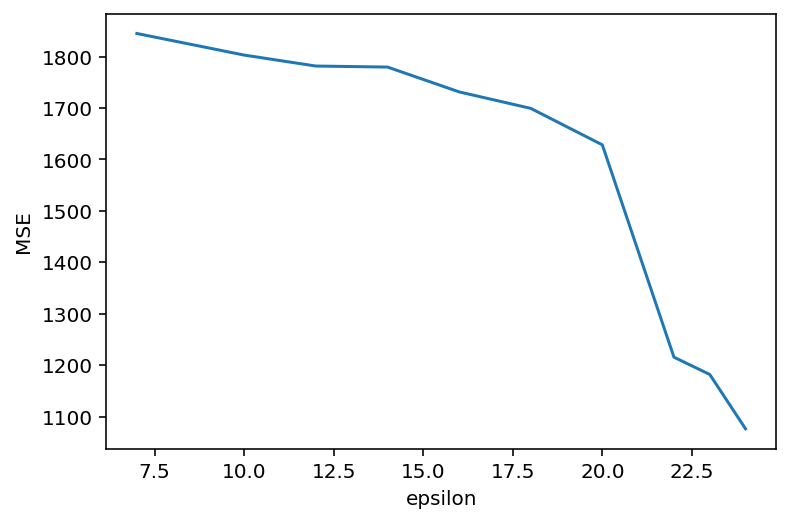}
    \caption{$\eps$ versus MSE for Criteo Sponsored Search Dataset}
    \label{fig:eps_opt_Criteo}
\end{minipage}

\end{figure}

For US Census, Figure \ref{fig:eps_opt} shows the mse loss incurred by the optimizer of the aggregated dataset on the original dataset using $\eps \in [14]$ and $\delta=1e-8$. As expected, the loss decreases as $\eps$ increases, stabalizing around $64.22$. All experimental numbers are averaged over 25 independent trials. Figure \ref{fig:eps_opt_Criteo} shows similar trends for the Criteo dataset using $\eps \in [7.0, 24.0]$, with the loss decreasing sharply to around $1076$ at $\eps = 24.0$.

In Appendix \ref{sec:additional_exp} we include empirical evaluations of  \LBA and \LLP for linear and neural regression respectively on two large-scale regression datasets, demonstrating that their utility preservation for several choices of bag size and number of sampled bags.

All experiments can be done on a single 16GB GPU and an 8-Core CPU with 128 GB of RAM. The code for the experiments will be publicly released along with the final version of the paper.

\section{Conclusions}\label{sec:Conclusions}
Our work shows that it is possible to achieve label-DP using enhancements of natural aggregation mechanisms with little or no additive label noise.
 Under a mild non-degeneracy conditions on the underlying data, \LFA and \LBA are label-DP while preserving the utility (optimum) of the linear regression task. Similar guarantees are provided by \LLP which ads additive noise to only a small fraction of the instance labels, while preserving the utility of general neural regression. 

Our experiments confirm that \LFA effectively preserves utility in linear regression, even under modest privacy budgets. We show that our results provide better privacy guarantees  compared to previous work which studied such mechanisms for differential privacy in linear regression.  Furthermore, our empirical evaluations of \LBA and \LLP for linear and neural regression tasks respectively demonstrate that they maintain utility even with a limited number of bags.

\bibliographystyle{abbrv}
\bibliography{references}

\appendix

\section{Some further preliminaries}
\begin{theorem}[Hoeffding's Bound, Thm 2. and Sec. 6 of \citep{Hoeffding}] \label{thm:Hoeffing}
If $X_1, \dots, X_n$ are sampled without replacement from some population $a_1, \dots, a_r \geq 0$, s.t. $\ol{X} = (1/n)\sum_{i=1}^n X_i$ and $\mu = \E[\ol{X}]$, then for $t > 0$, $\Pr[|\ol{X} - \mu| > t] \leq \tn{exp}\left(-2n^2t^2/\left(\sum_{i=1}^n a_i^2\right)\right)$.
\end{theorem}

The Hanson-Wright inequality (Theorem \ref{thm:hanson-wright}) implies the following concentration.
\begin{lemma}\label{lem:Gaussian-conc}
     For iid $X_i \sim N(0, 1)$  and $a_i \in \R$ ($i \in [n]$), let $S := \sum_{i=1}^n (a_iX_i)^2$ so that $\E[S] = \sum_{i=1}^n a_i^2$. Then,
    $\Pr\left[\left|S - \E[S]\right| > t \right] \leq 2\cdot\tn{exp}\left(-c_0\tn{min}\left(\frac{t^2}{\|\bv\|_2^2}, \frac{t}{\|\bv\|_\infty} \right)\right)$,
where $\bv = (a_1^2, \dots, a_n^2)$, and $c_0 > 0.145$ is an absolute constant (see \citep{moshksar2021absolute} for the lower bound on $c_0$).
\end{lemma}

Lemma F.3 in \citep{arbas2023polynomial} implies the following bound on the deviation between two perturbed Gaussians.

\begin{lemma}\label{lem:Gaussian-deviation} 
    Let $\bX_1 \sim N_d(\bm{\mu}_1, \sigma_1^2\bm{I}_d)$ and $\bX_2 \sim N_d(\bm{\mu}_2, \sigma_2^2\bm{I}_d)$ for $\bm{\mu}_1, \bm{\mu}_2 \in \R^d$ and $\sigma_1, \sigma_2 > 0$. If (i) $\max\left\{\left(\tfrac{\sigma_1}{\sigma_2}\right)^2, \left(\tfrac{\sigma_2}{\sigma_1}\right)^2\right\} \leq 1 + \tfrac{\kappa}{\sqrt{d}}$ and (ii) $\max\left\{\tfrac{\|\bm{\mu}_1 - \bm{\mu}_2\|_2}{\sigma_1}, \tfrac{\|\bm{\mu}_1 - \bm{\mu}_2\|_2}{\sigma_2}\right\} \leq \kappa$ for some $\kappa \in (0, 1/2]$ then for any $S \subseteq \R^d$, $\Pr[\bX_1 \in S] \leq e^\eps\Pr[\bX_1 \in S] + \delta$ for any $\eps \geq \kappa^2 + (2 + 2\sqrt{2})\kappa\sqrt{\log(2/\delta)} + 2\kappa\log(2/\delta)$.
\end{lemma}

\section{The \LLP mechanism and results} \label{sec:LLP-mechanism-results}
In this section we define the \LLP mechanism and state our privacy and utility bounds for it. Before that, we require some definitions in addition to those in Sec. \ref{sec:Ourresults}.
We also define $\mc{F}_N$ as the class of all neural-networks $f(\bx; \bs)$ characterized by some $\bs \in \R^d$ such that (i) $\|\bs\|_2 \leq B_{\sf s}$ for some $B_{\sf s} > 0$ and (ii) they follow the following lipschitzness property: $|f(\bx; \bs) - f(\bx; \ol{\bs})| \leq L\|\bs - \ol{\bs}\|_2$.

\begin{algorithm}[]
  \caption{Algorithm of \LLP}
  \label{fig-alg-Agg-Lin-LLP}
  \scriptsize
\begin{algorithmic}
  \STATE {\bfseries Input:} $\{y^{(i)}\}_{i=1}^n$ and $k, m \in \mathbb{Z}^+$ s.t. $n \gg mk$, $\mc{N} \subseteq [n]$. 
  \STATE Independently for each $i \in \mc{N}$, sample $g_i\sim N(0,1)$.
  \STATE Define: $tilde{y}^{(i)} = y^{(i)} + g_i  \tn{ if } i \in \mc{N} \tn{ and } y^{(i)} \tn{ otherwise.}$
  \STATE Define intermediate dataset $\tilde{\mb{D}} := \{\bx^{(i)}, \tilde{y}^{(i)}\}$.
  \STATE Sample $m$ Randomly sample disjoint subsets $S_1, \dots, S_m$ of $[n]$ s.t. $|S_j| = k$ and $S_j = \{i_{j1}, \dots, i_{jr}\}$, $j \in [m]$.
  \STATE For each $j \in [m]$ sample i.i.d. $w_{jr} \sim N(0, 1), r \in [k]$ and define $y_{\sf AGG}^{(j)} = \sum_{r = 1}^k w_{jr}\tilde{y}^{(i_{jr})}$ and $\bw_j = [w_{jr}]_{r=1}^k$.
  \STATE {\bfseries Output:} $\tilde{\mb{D}}_{\tn{LLP}} := \left\{\{(i_{jr}, \bx^{(i_{jr})}, j, w_{jr})\}, y_{\sf AGG}^{(j)}\right\}_{j=1}^m$.
\end{algorithmic}
\end{algorithm}

We consider \LLP defined in Algorithm \ref{fig-alg-Agg-Lin-LLP}. Essentially, for parameters $k$ and $m$ and any subset $\mc{N} \subseteq [n]$, \LLP first perturbs the labels corresponding to $\mc{N}$ using i.i.d. $N(0,1)$ additive noise, yielding an intermediate dataset. Next, it randomly samples $m$ disjoint $k$-sized subsets of $[n]$, and i.i.d. $N(0,1)$  multiplicative weights $w_{ij}$ for the indices in those subsets. Using these weights, for each subset it outputs the weighted sum of the labels corresponding to the indices in that subset, along with the weights themselves and the feature-vectors of that subset. Unlike \LBA, here the feature-vectors are not aggregated.

We use $B_{\tn{loss}}$ as an upper bound on squared loss on any point of $\mb{D}$ (which we explicitly calculate in Appendix \ref{subsec:llp_utlilty_prelim}). The output of the algorithm is a LLP dataset. Given access to it, the objective is to minimize $\val_{\sf LLP}(\tilde{\mb{D}}_{\tn{LLP}}, f) = \sum_{j=1}^m(y_{\sf AGG}^{(j)} - \sum_{i \in S_j}w_{ij}f(\bx^{(i)}))^2$. We prove (in Appendices \ref{sec:privacy-noisy_wtd_llp}, \ref{sec:neural-net-noisy_wtd_llp} and \ref{sec:utility-noisy_wtd_llp}) the following privacy and utility bounds.
\begin{theorem}\label{thm-intro-main3}
    For any $\rho \in (0, 1), \eps > 0$ and small $\theta$, the output $\tilde{\mb{D}}_{\tn{LLP}}$ of \LLP for the dataset $\mb{D}$ is $(\eps, \delta)$-label-DP whenever $k \geq O(B_{\sf Y}^2/\rho\eps^2 + B_{\sf Y}^4/\rho^2\eps^4)$ with
    \begin{align*}
        \delta = & \exp\Big{(}-\min\big{\{}\Omega\left(\tfrac{\eps^2\rho\sqrt{k}}{B_{\sf Y}^2}\right), \nonumber \\ 
        & \Omega(\rho^2k) - O(\log m), \Omega(\sqrt{k}) - O(\log m)\big{\}}\Big{)} \nonumber
    \end{align*}
    If $\tilde{\mb{D}}$ is a dataset with $N(0, 1)$ noise added to labels in $\mc{N}$ and $f^* = {\sf argmin}_{f \in \mc{F}_N}\val_{\sf LLP}(\tilde{\mb{D}}_{\sf LLP}, f)$ then $\val(\tilde{\mb{D}}, f^*) \leq \opt(\tilde{\mb{D}}, \mc{F}_1) + \theta$  except with probability at most
    \begin{align*}
       \exp\Big{(}-\min& \big{\{}\Omega(\tfrac{m\theta^2}{B_{\tn{loss}}^2}), \Omega(\tfrac{ mk\theta^2}{\rho B_{\tn{loss}}}), \Omega(m^{1/4}) - O(\log k), \nonumber \\
      & \Omega(\tfrac{\theta^2\sqrt{m}}{k^2 + kB_{\tn{loss}}})\big{\}} + O\left(d' \log\left(\tfrac{mLB_{\sf s}B_{\tn{loss}}}{\theta}\right)\right)\Big{)} \nonumber
    \end{align*}
    Fixing $B_{\sf Y}, B_{\sf s}, B_{\tn{loss}}, \rho, \eps, \theta, L$ as positive constants, we get $\delta = \exp\left(-\Omega(\sqrt{k}) + O(\log m)\right)$ and the failure probability of preservation of regression optimum as $\exp(-\min\left\{\Omega(m^{1/4}) - O(\log m), \Omega\left(\sqrt{m}/k^2\right)\right\} +$
    $O(d'\log m))$.
\end{theorem}

Theorem \ref{thm-intro-main3} shows label-DP guarantees by perturbing \emph{any} $\rho > 0$ fraction of the labels followed by label aggregation, making it efficient in the additive noise. Furthermore, since the feature-vectors in bags along with their weights are available, one can perform neural mse-regression whose utility is similarly preserved.
The proof uses the Hoeffding bound to show that for a random $k$-size bag, $\approx \rho$-fraction of its points are noisy w.h.p. This additive noise along with appropriate tail bounds on the magnitude of the aggregation weights for this bag along with Gaussian deviation bounds yields the desired privacy guarantee.   Appendix \ref{sec:privacy-noisy_wtd_llp} contains the detailed proof.

The utility bound -- proved in Appendix \ref{sec:utility-noisy_wtd_llp}  -- is similar to that of \LBA, except that the errors are additive to the addition of additive noise.

\section{Limited Privacy of Naive \LBA and \LLP}\label{sec:no-DP-example}
Consider a dataset $\mb{D}$ as defined in Section \ref{sec:Ourresults}, and the descriptions of \LBA and \LLP therein. In the naive variant of \LBA there are no random aggregation weights $w_i$ and the output is  $\left\{S_j, (\bx_{\sf AGG}^{(j)}, y_{\sf AGG}^{(j)}) := \sum_{i\in S_j} (\bx^{(i)}, y^{(i)})\right\}_{j=1}^m$. Consider a change at $y^{(i^*)}$ for some $i^* \in [n]$ to yield $\mb{D}'$. Now, if $i^* \in S_{j^*}$ for some $j^* \in [m]$, then $y_{\sf AGG}^{(j^*)}$ is also changed and therefore no output of the algorithm with input $\mb{D}$ is possible when the input is $\mb{D}'$. Since $\Pr\left[\exists j^* \in [m]\tn{ s.t. } i^* \in S_{j^*}\right] = mk/n$, this mechanism cannot be $(\eps, \delta)$-label-DP with $\delta < mk/n$ for any $\eps > 0$. This is unsatisfactory as $mk/n$ increases with both the bag size and number of bags, and in many applications could be a significant fraction.

In naive \LLP, there are neither any random aggregation weights nor any random additive noise $g_i$, and the output is $\left\{\{(i, j, \bx^{(i)})\,\mid\, i \in S_j\}, y_{\sf AGG}^{(j)} := \sum_{i\in S_j}y^{(i)}\right\}_{j=1}^m$. The same argument holds: a change in  $y^{(i^*)}$ leads to a change in  $y_{\sf AGG}^{(j^*)}$ as long as  $i^* \in S_{j^*}$ for some $j^* \in [m]$. Thus, the mechanism cannot be $(\eps, \delta)$-label-DP with $\delta < mk/n$ for any $\eps > 0$.

\section{The case of $\mb{y}$ in $\tn{colspace}(\mb{X})$ for \LFA} \label{sec:yincolspaceX}
Let $\bX$ be the matrix whose rows are the feature vectors of the dataset and $\by$ be the vector given by the labels. Let $\bw$ be the weight vector sampled for generating one aggregate yielding $\bx_{\sf AGG}^{\sf T} := \bw^{\sf T}\bX$ and $\bw^{\sf T}\by$ as the output. Suppose for the moment that $\by$ belongs to the column-space of $\bX$, i.e. there is $\mb{h}$ s.t. $\by = \bX\mb{h}$.

Define the following quantities: 
\begin{itemize}
    \item Perturbed label vector $\tilde{\by} = \by + t\mb{e}_i$ where $\mb{e}_i$ is the $i^{\sf th}$ coordinate vector for some $i \in [n]$ and $t \in \R\setminus\{0\}$.
    \item $A := \{(\ba^{\sf T}\bX, \ba^{\sf T}\by)\,\mid\, \ba \in \R^n\}$ and $B = \{(\bb^{\sf T}\bX, \bb^{\sf T}\tilde{\by})\,\mid\, \bb \in \R^n\}$ as the sets of all possible outputs of \LFA with label vectors $\by$ and $\tilde{\by}$ respectively with $m = 1$.
    \item $\mc{H}$ as the distribution on $(\ba^{\sf T}\bX, \ba^{\sf T}\by)$ induced by $\ba \in N(0,1)^n$, and $\tilde{\mc{H}}$ as the distribution on $(\bb^{\sf T}\bX, \bb^{\sf T}\tilde{\by})$ induced by $\bb \sim N(0,1)^n$. In particular, $\mc{H}$ and $\tilde{\mc{H}}$ are the distributions over the outputs of \LFA with label vectors $\by$ and $\tilde{\by}$ respectively with $m = 1$.
\end{itemize}

Observe that 
\begin{equation}
    A \cap B = \{(\bb^{\sf T}\bX, \bb^{\sf T}\by)\,\mid\, \bb \in \R^n \tn{ s.t. } \bb^{\sf T}\bX =  \ba^{\sf T}\bX \tn{ and }  \bb^{\sf T}\by = \ba^{\sf T}\tilde{\by} \tn{ for some } \ba \in \R^n\}. \nonumber\label{eqn:AcapB}
\end{equation}
Suppose $\ba,\bb \in \R^n$ satisfy $\bb^{\sf T}\bX =  \ba^{\sf T}\bX \tn{ and }  \bb^{\sf T}\tilde{\by} = \ba^{\sf T}\by$. Let $\bz = \ba^{\sf T}\bX = \bb^{\sf T}\bX$ for some row vector $\bz$. Then, since $\by = \bX\mb{h}$, we have $\ba^{\sf T}\by = \ba^{\sf T}\bX\mb{h} = \bz\mb{h}$. Also, $\bb^{\sf T}\tilde{\by} = \bb^{\sf T}(\by + t\mb{e}_i) = \bb^{\sf T}\by +  t\bb^{\sf T}\mb{e}_i = \bb^{\sf T}\bX\mb{h} + tb_i = \bz\mb{h} + tb_i$, where $b_i$ is the $i^{\sf th}$ coordinate of $\bb$. However, since $\bb^{\sf T}\tilde{\by} = \ba^{\sf T}\tilde{\by}$ we obtain that $tb_i = 0 \Rightarrow b_i = 0$ since $t \neq 0$. Therefore, we obtain that
\begin{equation}
    A \cap B \subseteq \{(\bb^{\sf T}\bX, \bb^{\sf T}\by)\,\mid\, \bb \in \R^n, b_i = 0\} \Rightarrow \Pr_{\tilde{\mc{H}}}[A] = \Pr_{\tilde{\mc{H}}}[A\cap B] \leq \Pr_{b_i \leftarrow N(0,1)}[b_i = 0] = 0, \nonumber
\end{equation}
where we we have $\Pr_{\tilde{\mc{H}}}[A] = \Pr_{\tilde{\mc{H}}}[A\cap B]$ since $B$ is the support of $\tilde{\mc{H}}$. On the other hand, $A$ is the support of $\mc{H}$, therefore $\Pr_{\mc{H}}[A] = 1$. Therefore, the set $A$ satisfies $\Pr_{\mc{H}}[A] = 1$ as well as $\Pr_{\tilde{\mc{H}}}[A] = 0$ which violates the label-DP condition (Defn. \ref{def:label-DP}) for any $\eps$ unless $\delta = 1$ i.e., there is no privacy guarantee.

\section{Covering Number of a Ball}
\begin{lemma}\label{lem:covering_no_ball}
The $\eps$-covering number of the radius-$K$ Euclidean ball $\mathbb{B}^d_2(K)$ and the radius-$K$ Euclidean sphere $K\mathbb{S}^{d-1}$ is at most $\left(1 + \tfrac{2K}{\eps}\right)^d$.
\end{lemma}
\begin{proof}
Using Corollary 4.2.13 of \citep{Vershynin-book}, we know that the $\eps'$-covering number of $\mathbb{S}^{d-1}$ and  $\mathbb{B}^d_2(1)$ is at most $(1 + 2/\eps')^d$. Choosing $\eps' = \eps/K$ and scaling by a factor of $K$ yields $\mathbb{B}^d_2(K)$ and  $K\mathbb{S}^{d-1}$ along with the respective $\eps$-nets.
\end{proof}

\section{Matrix Chernoff Bound for sampling without replacement}\label{appendix:matrix_chernoff_wo_replacement}
In this section we prove the Matrix Chernoff bound for sampling matrices uniformly at random without replacement. We will use this in Appendix \ref{sec:privacynet}.
\begin{lemma}\label{lem:subadditivity_without_replacement}
Consider a finite sequence $\{\bX_i\}_{i=1}^k$ of Hermitian matrices sampled uniformly at random without replacement from a finite set of matrices of dimension $d$. Then,
\begin{align}
    \E\left[{\sf tr}\exp\left(\sum_{i=1}^k\theta\bX_i\right)\right] \leq {\sf tr}\exp\left(\sum_{i=1}^k\log\E e^{\theta\bX_i}\right) \nonumber
\end{align}
\end{lemma}
\begin{proof}
Let $\bY_i$ be sampled uniformly at random with replacement from the same set. Then using the result in \citep{notesamplingwihtoutreplacement}, we have
\begin{align}
    \forall \theta \in \R, \qquad \E\left[{\sf tr}\exp\left(\sum_{i=1}^k\theta\bX_i\right)\right] \leq \E\left[{\sf tr}\exp\left(\sum_{i=1}^k\theta\bY_i\right)\right] \nonumber
\end{align}
Hence, for all $\theta \in \R$
\begin{align}
    \E\left[{\sf tr}\exp\left(\sum_{i=1}^k\theta\bX_i\right)\right] & \leq \E\left[{\sf tr}\exp\left(\sum_{i=1}^k\theta\bY_i\right)\right]  \leq {\sf tr}\exp\left(\sum_{i=1}^k\log\E e^{\theta\bY_i}\right)  = {\sf tr}\exp\left(\sum_{i=1}^k\log\E e^{\theta\bX_i}\right) \nonumber
\end{align}
where the second inequality comes from subadditivity of matrix cgfs when self-adjoint matrices are independently sampled (Lemma 3.4 in \citep{randommatrixbounds}) and the equality comes from the fact that $\E e^{\theta\bY_i} = \E e^{\theta\bX_i}$.
\end{proof}
\begin{lemma}[The laplace transform method (Proposition 3.1 in \citep{randommatrixbounds})]\label{lem:laplace_transform_method}
Let $\bY$ be a random self-adjoint matrix. For all $t \in \R$,
\begin{align}
    \Pr[\lambda_{\max}(\bY) \geq t] \leq \inf_{\theta > 0}\{e^{-\theta t}\cdot\E{\sf tr}e^{\theta \bY}\} \nonumber
\end{align}
\end{lemma}
\begin{proof}
Look at the proof of Proposition 3.1 in \citep{randommatrixbounds}.
\end{proof}
\begin{lemma}\label{lem:master_tail_bound}
Consider a finite sequence $\{\bX_i\}_{i=1}^k$ of Hermitian matrices sampled uniformly at random without replacement from a fixed set. Then for all $t \in \R$,
\begin{align}
\Pr\left[\lambda_{\max}\left(\sum_{i=1}^k\bX_i\right) \geq t\right] \leq \inf_{\theta > 0}\left\{e^{-\theta t}{\sf tr}\exp\left(\sum_{i=1}^k\log \E e^{\theta X_i}\right)\right\} \nonumber
\end{align}
\end{lemma}
\begin{proof}
Since $\sum_{i=1}^k\bX_i$ is a Hermitian random matrix, by Lemma \ref{lem:laplace_transform_method},
\begin{align}
    \Pr\left[\lambda_{\max}\left(\sum_{i=1}^k\bX_i\right) \geq t\right] \leq \inf_{\theta > 0}\left\{e^{-\theta t}\cdot\E\left[{\sf tr}\exp\left(\sum_{i=1}^k\theta \bX_i\right)\right]\right\} \nonumber
\end{align}
Using the inequality in Lemma \ref{lem:subadditivity_without_replacement}, since $\{\bX_i\}_{i=1}^k$ is a finite sequence of Hermitian matrices, we obtain the desired inequality.
\end{proof}
\begin{lemma}\label{lem:pre_bound}
Consider a finite sequence $\{\bX_i\}_{i=1}^k$ of Hermitian matrices of dimension $d$ sampled uniformly at random without replacement from a fixed set. Then for all $t \in \R$,
\begin{align}
    \Pr\left[\lambda_{\max}\left(\sum_{i=1}^k\bX_i\right) \geq t\right] \leq d \cdot \inf_{\theta > 0}\exp\left(-\theta t + n \cdot \log \lambda_{\max}\left(\tfrac{1}{n}\sum_{i=1}^k \E e^{\theta \bX_i}\right)\right) \nonumber
\end{align}
\end{lemma}
\begin{proof}
The matrix logarithm operator is concave \citep{randommatrixbounds}. Thus, for all $\theta > 0$,
\begin{align}
    \sum_{i=1}^k\log \E e^{\theta \bX_i} = n \cdot \tfrac{1}{n}\sum_{i=1}^k\log \E e^{\theta \bX_i} \leq n \cdot \log\left(\tfrac{1}{n}\sum_{i=1}^k\log \E e^{\theta \bX_i}\right)\label{eqn:log_ineq}
\end{align}
Now, trace exponential is a monotone \citep{randommatrixbounds}. Using Lemma \ref{lem:master_tail_bound} and equation \eqref{eqn:log_ineq} we get that
\begin{align}
    \Pr\left[\lambda_{\max}\left(\sum_{i=1}^k\bX_i\right) \geq t\right] \leq e^{-\theta t} {\sf tr}\exp\left(n \cdot \log\left(\tfrac{1}{n}\sum_{i=1}^k\log \E e^{\theta \bX_i}\right)\right) \nonumber
\end{align}
We next bound the trace by $d$ times the maximum eigenvalue.
\begin{align}
    \Pr\left[\lambda_{\max}\left(\sum_{i=1}^k\bX_i\right) \geq t\right] \leq d e^{-\theta t} \lambda_{\max}\left(\exp\left(n \cdot \log\left(\tfrac{1}{n}\sum_{i=1}^k\log \E e^{\theta \bX_i}\right)\right)\right) \nonumber
\end{align}
Using the spectral theorem twice, we get the desired result.
\end{proof}
\begin{lemma}\label{lem:matrix_chernoff}
Consider a finite sequence $\{\bX_i\}_{i=1}^k$ of Hermitian matrices of dimension $d$ sampled uniformly at random without replacement from a fixed set. Let \begin{align}
    \bX_i \geq 0 \qquad \lambda_{\max}(\bX_i) \leq R \nonumber
\end{align}
Define $\mu_{\min} := \lambda_{\min}\left(\sum_{i = 1}^k\E \bX_i\right)$. Then,
\begin{align}
    \Pr\left[\lambda_{\min}\left(\sum_{i=1}^k\bX_i\right) \leq (1 - \delta)\mu_{\min}\right] \leq d \cdot \left[\frac{e^{-\delta}}{(1 - \delta)^{1 - \delta}}\right]^{\mu_{\min}/R}\nonumber
\end{align}
\end{lemma}
\begin{proof}
Lemma 5.8 in \citep{randommatrixbounds} states that for any random psd matrix $\bX$ such that $\lambda_{\max}(\bX) \leq 1$,
\begin{align}\label{eqn:chernoff_mgf}
    \E[e^{\theta \bX}] \leq \mb{I} + (e^\theta - 1)(\E \bX) \qquad \forall \theta \in \R 
\end{align}
In order to apply lemma \ref{lem:pre_bound} on the sequence $\{-\bX_i/R\}_{i=1}^n$, we use \eqref{eqn:chernoff_mgf} as below
\begin{align}
    \E e^{\theta(-\bX_i/R)} = \E e^{(-\theta)\bX_i/R} \leq \mb{I} - (g(\theta)/R)(\E \bX_i) \qquad g(\theta) := 1 - e^{-\theta} \nonumber
\end{align}
Note that $\lambda_{\min}(-\mb{A}) = -\lambda_{\max}(\mb{A})$ for any $\mb{A} \in \R^{d \times d}$. Thus, for any $t \in \R$, using Lemma \ref{lem:pre_bound}
\begin{align}
    \Pr\left[\lambda_{\min}\left(\sum_{i=1}^k\bX_i\right) \leq t\right] & = \Pr\left[\lambda_{\max}\left(\sum_{i=1}^k(-\bX_i/R)\right) \geq -t/R\right] \nonumber \\
    & \leq d \cdot \exp\left(\theta t/R + n \cdot \log \lambda_{\max}\left(\tfrac{1}{n}\sum_{i=1}^k\left(\mb{I} - g(\theta) \E \bX_i/R\right)\right)\right) \nonumber \\
    & = d \cdot \exp\left(\theta t/R + n \cdot \log \left( 1 - (g(\theta)/R) \lambda_{\max}\left(\tfrac{1}{n}\sum_{i=1}^k\E \bX_i\right)\right)\right) \nonumber \\
    & = d \cdot \exp\left(\theta t/R + n \cdot \log \left( 1 - g(\theta)\mu_{\min}/nR \right)\right) \nonumber \\
    & \leq d \cdot \exp\left(\theta t/R -g(\theta)\mu_{\min}/R \right) \qquad [\because g(\theta)\mu_{\min}/nR \leq 1] \nonumber 
\end{align}
Choose $t \rightarrow (1 - \delta)\mu_{\min}$ and $\theta = - \log(1 - \delta)$ to get the result.
\end{proof}

\section{Utility analysis of \LFA}\label{appendix:lfa_utility}
For the rest of the section, define $B_{\sf l} = B_{\sf Y} + B_{\sf X}B_{\sf R}$. It can be seen using the triangular inequality and the cautchy-schwartz inequality that $B_{\sf l}^2$ is an upper bound on the mse loss that any linear regressor in $\mc{F}_l$ can incur on an example in a dataset. We first prove that the loss incurred by any linear regressor on the aggregated dataset stays within a multiplicative factor of the loss incurred on the original dataset with high probability.
\begin{lemma}\label{lem:loss_multiplicative_bound}
Let $\mb{D}_{\sf AGG} = \{\bx_{\sf AGG}^{(j)}, y_{\sf AGG}^{(j)}\}_{j=1}^m$ be the output of Algorithm \ref{alg:bag_aggregation} when it is given the input $\mb{D} = \{\bx^{(i)}, y^{(i)}\}_{i=1}^n$ and the number of bags as $m$. For any $\br \in \R^d$ and any $\theta \in (0, 1)$ w.p. at least $1 - 2\exp(-c_0 m \theta^2)$,
\begin{align}
    \left|\tfrac{1}{m}\val(\mb{D}_{\sf AGG}, \br^{\sf T}\bx) - \val(\mb{D}, \br^{\sf T}\bx)\right| \leq \theta \cdot \val(\mb{D}, \br^{\sf T}\bx) \nonumber
\end{align}
\end{lemma}
\begin{proof}
Let $\val(\mb{D}, \br^{\sf T}\bx) = \omega n$ and define $G_j := \sum_{i=1}^n w_{ij}\left(y^{(i)} - \br^{\sf T}\bx^{(i)}\right)$. Thus, the loss incurred on $j^{\sf th}$ example of $\mb{D}_{\sf AGG}$ is $G_j^2$. Note that $G_j$ is a mean-zero gaussian with $\Ex[G_j^2] = \sum_{i=1}^n \left(y^{(i)} - \br^{\sf T}\bx^{(i)}\right)^2 = \omega n$. We apply lemma \ref{lem:Gaussian-conc} as follows: $\bv = \left(\Ex[G_j^2]\right)_{j=1}^m$, so that $\|\bv\|_2^2 = m\omega^2n^2$ and $\|\bv\|_\infty = \omega n$. Thus, $\Pr[|\sum_{j=1}^m G_j^2 - \sum_{j=1}^m \Ex[G_j^2]| \geq \theta nm\omega] \leq 2\exp\left(-c_0\min\left\{\tfrac{(\theta nm\omega)^2}{m\omega^2n^2}, \tfrac{\theta nm\omega}{\omega n}\right\}\right) = 2\exp(-c_0\theta^2m)$. 
\end{proof}
To prove that the minima is preserved, we need to claim that the event described in the statement of Lemma \ref{lem:loss_multiplicative_bound} occurs for all linear regressors on $\mc{F}_l$. We do this by constructing a net over $\mc{F}_l$ and taking a union bound over this event happening for all linear regressors in that net. We then demonstrate the existence of a multiplicative error bound between the loss (on the original as well as the aggregated dataset) incurred by a linear regressor and the linear regressor closest to it in the net.
\begin{lemma}\label{lem:instance_loss_net_<name>}
Given a dataset $\mb{D} = \{\bx^{(i)}, y^{(i)}\}_{i=1}^n$ such that $\inf_{\br \in \mathbb{B}^d_{2}(B_{\sf R})}\left(y^{(i)} - \br^{\sf T}\bx^{(i)}\right)^2 \geq \gamma n$, if $\ol{\br}$ in an $(\nu \gamma/2B_{\sf l}B_{\sf X})$-net over $\mathbb{B}^d_{2}(B_{\sf R})$ is the closest to $\br \in \mathbb{B}^d_{2}(B_{\sf R})$ then $\left|\val(\mb{D}, \ol{\br}^{\sf T}\bx) - \val(\mb{D}, \br^{\sf T}\bx)\right| \leq \nu \cdot \val(\mb{D}, \br^{\sf T}\bx)$.
\end{lemma}
\begin{proof} We compute $\left|\val(\mb{D}, \ol{\br}^{\sf T}\bx) - \val(\mb{D}, \br^{\sf T}\bx)\right|$
\begin{align}
    & = \left|\sum_{i=1}^n\left(y^{(i)} - \br^{\sf T}\bx^{(i)}\right)^2 - \sum_{i=1}^n\left(y^{(i)} - \ol{\br}^{\sf T}\bx^{(i)}\right)^2\right| \nonumber \\
    & \leq 2B\sum_{i=1}^n\left|(\br - \ol{\br})^{\sf T}\bx^{(i)}\right| \leq 2B\sum_{i=1}^n\|\br - \ol{\br}\|_2\|\bx^{(i)}\|_2 \nonumber \\
    & \leq \nu\gamma n \leq \nu \cdot \val(\mb{D}, \br^{\sf T}\bx) \nonumber
\end{align}
\end{proof}
\begin{lemma}\label{lem:bag_loss_net_<name>}
Let a dataset $\mb{D}_{\sf AGG} = \{\bx_{\sf AGG}^{(j)}, y_{\sf AGG}^{(j)}\}_{j=1}^m$ be the output of Algorithm \ref{alg:bag_aggregation} when it recieves a dataset of size $n$ and number of aggregates $m \in \mathbb{N}$ as input. Let $\mc{T}$ be a $(\nu \gamma/2m^2nB_{\sf l}B_{\sf X})$-net over $\mathbb{B}^d_{2}(B_{\sf R})$. If $\inf_{\br \in \mc{T}}\val(\mb{D}_{\sf AGG}, \br^{\sf T}\bx) \geq \gamma mn$ and if $\ol{\br}$ is the projection of $\br \in \mathbb{B}^d_{2}(B_{\sf R})$ on $\mc{T}$ then $\left|\val(\mb{D}_{\sf AGG}, \br^{\sf T}\bx) - \val(\mb{D}_{\sf AGG}, \ol{\br}^{\sf T}\bx)\right| \leq \nu\cdot\val(\mb{D}_{\sf AGG}, \ol{\br}^{\sf T}\bx)$ with probability at least $1 - n\cdot\sqrt{2/\pi}\exp(-m^2/2)$.
\end{lemma}
\begin{proof} We evaluate $\left|\val(\mb{D}_{\sf AGG}, \ol{\br}^{\sf T}\bx) - \val(\mb{D}_{\sf AGG}, \br^{\sf T}\bx)\right|$ as
\begin{align}
    & = \left|\sum_{j=1}^m\left(y_{\sf AGG}^{(j)} - \br^{\sf T}\bx_{\sf AGG}^{(j)}\right)^2 - \sum_{j=1}^m\left(y_{\sf AGG}^{(j)} - \ol{\br}^{\sf T}\bx_{\sf AGG}^{(j)}\right)^2\right| \nonumber \\
    & \leq \sum_{j=1}^m\left|2y_{\sf AGG}^{(j)} - (\br + \ol{\br})^{\sf T}\bx_{\sf AGG}^{(j)}\right|\|\br - \ol{\br}\|_2\|\bx_{\sf AGG}^{(j)}\|_2 \label{eq:d_agg_loss_net}
\end{align}
Now, a standard gaussian $g$ satisfies
\begin{align}
    \Pr[|g| \leq m] \geq 1 - \tfrac{1}{m}\sqrt{\tfrac{2}{\pi}}\exp\left(-\tfrac{m^2}{2}\right) \nonumber
\end{align}
Taking a union bound over all gaussian weights $w_{ij}$,
\begin{align}
    \Pr[\wedge_{i=1}^n\wedge_{j=1}^m|w_{ij}| \leq m] \geq 1 - n\sqrt{\tfrac{2}{\pi}}\exp\left(-\tfrac{m^2}{2}\right) \nonumber
\end{align}
If all the weights are bounded by $m$, then $\|\bx_{\sf AGG}^{(j)}\| \leq nmB_{\sf X}$ and $\left|2y_{\sf AGG}^{(j)} - (\br + \ol{\br})^{\sf T}\bx_{\sf AGG}^{(j)}\right| \leq 2nmB_{\sf l}$. We use equation \eqref{eq:d_agg_loss_net} to obtain that: $\left|\val(\mb{D}_{\sf AGG}, \ol{\br}^{\sf T}\bx) - \val(\mb{D}_{\sf AGG}, \br^{\sf T}\bx)\right| \leq m(2nmB_{\sf l})(\nu \gamma/2m^2nB_{\sf l}B_{\sf X})(nmB_{\sf X}) \leq \nu\gamma mn \leq \nu\cdot\val(\mb{D}_{\sf AGG}, \ol{\br}^{\sf T}\bx)$.
\end{proof}
Using the above lemmas, we complete the proof of the privacy guarantee as follows. Let $\mc{T}$ be a $(\nu\gamma/4m^2nB_{\sf l}B_{\sf X})$-net over $\mathbb{B}^d_{2}(B_{\sf R})$ for some $\nu > 0$. Using Lemma \ref{lem:covering_no_ball} and the fact that $B_{\sf R}B_{\sf X} \leq B_{\sf l}$, $|\mc{T}| \leq (12m^2nB_{\sf l}^2/\nu\gamma)^d$ whenever $\nu \leq (4m^2nB_{\sf l}B_{\sf X}/\gamma)$. Taking a union bound of the event in Lemma \ref{lem:loss_multiplicative_bound} over all linear regressors $\{\br^{\sf T}\bx\,|\,\br \in \mc{T}\}$, with probability at least $1 - \left(\tfrac{12m^2nB_{\sf l}^2}{\nu\gamma}\right)^d\exp(-\tfrac{c_0m\theta^2}{256})$, for all $\ol{\br} \in \mc{T}$,
\begin{align}
    \val(\bD_{\sf AGG}, \ol{\br}^{\sf T}\bx) \in \left(1 - \tfrac{\theta}{16}, 1 + \tfrac{\theta}{16}\right) \cdot m\val(\bD, \ol{\br}^{\sf T}\bx) \label{eq:loss_on_net_bound}
\end{align}
For any linear regressor $\br \in \mathbb{B}_2^d(B_{\sf R})$, let $\ol{\br} \in \mc{T}$ be the projection of $\br$ on $\mc{T}$. Using Lemma \ref{lem:instance_loss_net_<name>},
\begin{align}
    \val(\bD, \ol{\br}^{\sf T}\bx) \in \left(1 - \tfrac{\nu}{2m^2n}, 1 + \tfrac{\nu}{2m^2n}\right) \cdot \val(\bD, \br^{\sf T}\bx)\label{eq:loss_net_to_out_bound}
\end{align}
Also, using Lemma \ref{lem:bag_loss_net_<name>}, if the event in equation \eqref{eq:loss_on_net_bound} occurs with $\theta \leq 8$ then $\inf_{\br \in \mc{T}}\val(\bD_{\sf AGG}, \br^{\sf T}\bx) \geq \gamma nm/2$ and thus with probability at least $1 - n\sqrt{2/\pi}\exp(-m^2/2)$,
\begin{align}
    \val(\bD_{\sf AGG}, \br^{\sf T}\bx) \in \left(1 - \nu, 1 + \nu\right) \cdot \val(\bD_{\sf AGG}, \br^{\sf T}\bx)\label{eq:loss_out_to_net_bound}
\end{align}
Using equations \eqref{eq:loss_out_to_net_bound},  \eqref{eq:loss_on_net_bound} and \eqref{eq:loss_net_to_out_bound} in that order, we get that for all $\br \in \mathbb{B}_2^d(B_{\sf R})$,
\begin{align}
    \val(\bD_{\sf AGG}, \br^{\sf T}\bx) \in \left(a, b\right) \cdot m\val(\bD, \br^{\sf T}\bx) \nonumber
\end{align}
Where $a = (1-\theta/16)(1-\nu)(1-\nu/2m^2n)$ and $b = (1+\theta/16)(1+\nu)(1+\nu/2m^2n)$. Picking $\nu = \theta/16$, when $\theta \leq 8$ ($n, m \geq 1$) we obtain that
\begin{align}
    \val(\bD_{\sf AGG}, \br^{\sf T}\bx) \in \left(1-\theta/4, 1+\theta/4\right) \cdot m\val(\bD, \br^{\sf T}\bx) \nonumber
\end{align}
with probability at least $1 - n\sqrt{\tfrac{2}{\pi}}\exp\left(-\tfrac{m^2}{2}\right) - \left(\tfrac{192m^2nB_{\sf l}^2}{\theta\gamma}\right)^d\exp\left(-\tfrac{c_0m\theta^2}{256}\right)$. Given this event occurs, let $\br_{{\sf AGG}*} = {\sf argmin}_{\br \in \mathbb{B}_2^d(B_{\sf R})}\val(\bD_{\sf AGG}, \br^{\sf T}\bx)$ and $\br_{*} = {\sf argmin}_{\br \in \mathbb{B}_2^d(B_{\sf R})}\val(\bD, \br^{\sf T}\bx)$. Then, $\val(\bD, \br_{{\sf AGG}*}^{\sf T}\bx) \leq \tfrac{1}{m(1-\theta/4)}\val(\bD_{\sf AGG}, \br_{{\sf AGG}*}^{\sf T}\bx) \leq \tfrac{1}{m(1-\theta/4)}\val(\bD_{\sf AGG}, \br_*^{\sf T}\bx) \leq \tfrac{1 + \theta/4}{1-\theta/4}\opt(\bD, \mc{F}_l) \leq (1 + \theta)\opt(\bD, \mc{F}_l)$ whenever $\theta \leq 2$.

\section{Utility for \LBA}\label{sec:LBA-utility}
In the next set of lemmas in this section, we shall prove that $(1 - \theta)\tfrac{1}{n}\val(\mb{D}, f) \leq \tfrac{1}{mk}\val(\hat{\mb{D}}, f) \leq (1 + \theta)\tfrac{1}{n}\val(\mb{D}, f)$ for all $f \in \mc{F}_l$ with probability at least that mentioned in Theorem \ref{thm-intro-main1}. The rest of the proof follows from the following argument. Let $f^* = {\sf argmin}_{f \in \mc{F}_0}\val(\hat{\mb{D}}, f)$ and $f^{**} = {\sf argmin}_{f \in \mc{F}_0}\val(\mb{D}, f)$, then our result follows by the following argument.
\begin{align}
    (1 - \theta)\tfrac{1}{n}\val(\mb{D}, f^*) \leq \tfrac{1}{mk}\val(\hat{\mb{D}}, f^*) \leq \tfrac{1}{mk}\val(\hat{\mb{D}}, f^{**}) \leq (1 + \theta)\tfrac{1}{n}\val(\mb{D}, f^{**}) \nonumber
\end{align}
\begin{lemma}\label{lem:single_regressor_bound_general}
Let $\{\bx_{\sf AGG}^{(j)}, y_{\sf AGG}^{(j)}\}_{j=1}^m$ be returned by \LBA (Alg. \ref{fig-alg-Agg-Lin-LBA}). Then, for any linear regressor $\br^{\sf T}\bx$ s.t. $\|\br\|_2 \leq B_3$ and $\sum_{i=1}^n\left(y^{(i)} - \br^{\sf T}\bx^{(i)}\right)^2 = \omega n$, and any $\theta \in (0, 1)$ w.p. at least
$1 - 4\cdot\tn{exp}(-2mk\theta^2\omega^2/B_{\sf l}^4)$ we have $(1 - 2\theta)km\omega \leq \sum_{j=1}^m\left(y_{\sf AGG}^{(j)} - \br^{\sf T}\bx_{\sf AGG}^{(j)}\right)^2 \leq (1 + 2\theta)km\omega$. Note that $\omega \geq \gamma$ by the assumption in Theorem \ref{thm-intro-main1}.
\end{lemma}
\begin{proof}
Fix a bag $S_j = \{i_{j1}, ..., i_{jk}\}$ for some $j = 1, \dots, m$. Then, its aggregated feature vector $\bx_{\sf AGG} = \sum_{r=1}^k w_{jr}\bx^{(i_{jr})}$ and the aggregated label $y_{\sf AGG} = \sum_{r=1}^k w_{jr}y^{(i_{jr})}$. The corresponding loss is $G_j^2$ where  $G_j := \sum_{r=1}^kw_{jr}\left(y^{(i_{jr})} - \br^{\sf T}\bx^{(i_{jr})}\right)$, is a mean-zero Gaussian random variable s.t. $\E[G_j^2] = \sum_{r=1}^k\left(y^{(i_{jr})} - \br^{\sf T}\bx^{(i_{jr})}\right)^2$. Note that $\sum_{j=1}^m \left[\E[G_j^2]\right] = \sum_{j=1}^m\sum_{r=1}^k \left(y^{(i_{jr})} - \br^{\sf T}\bx^{(i_{jr})}\right)^2$ which is the sum of $mk$ samples without replacement from a population such that each sample is in $[0,B_{\sf l}^2]$. The expectation of this sum is $\E_{S_1, \dots, S_m}\sum_{j=1}^m\sum_{r=1}^k \left(y^{(i_{jr})} - \br^{\sf T}\bx^{(i_{jr})}\right)^2 = km\omega$. We can apply Theorem \ref{thm:Hoeffing} so that, $\Pr\left[\left|\sum_{j=1}^m \left[\E[G_j^2]\right] - km\omega\right| \geq \theta km\omega\right] \leq 2\cdot\exp\left(-2mk\theta^2\omega^2/B_{\sf l}^4\right)$.

Fix $S_1, \dots, S_m$ s.t. the condition inside the probability on the LHS of above holds. Observe that $\E[G_j^2] \leq B_{\sf l}^2k$. 
Using this we apply the bound from Lemma \ref{lem:Gaussian-conc} as follows: let $\bv = \left(\E[G_j^2]\right)_{j=1}^m$, so that $\|\bv\|_2^2 \leq B_{\sf l}^4mk^2$, and  $\|\bv\|_\infty \leq B_{\sf l}^2k$ yielding
$
    \Pr\left[\left|\sum_{j=1}^m G_j^2 - km\omega\right| \geq 2\theta km\omega \right] 
    \leq  \Pr\left[\left|\sum_{j=1}^m G_j^2 - \E\left[\sum_{j=1}^m G_j^2\right]\right| \geq \theta km\omega \right] \nonumber
    \leq 2\tn{exp}\left(-c_0\min\left\{\frac{(\theta km\omega)^2}{B_{\sf l}^4mk^2}, \frac{\theta km\omega}{B_{\sf l}^2k} \right\}\right) =  2\tn{exp}\left(-\frac{c_0m\theta^2\omega^2}{B_{\sf l}^4}\right), \nonumber
$
since $\theta \in (0,1)$ and $\omega \in (0, B_{\sf l}^2)$. 
Combining the above probabilistic bounds completes the proof.
\end{proof}

\begin{lemma}\label{lem:instance_general_to_net_general}
Let a dataset $\{\bx^{(i)}, y^{(i)}\}_{i=1}^n$ such that any linear regressor has an avearge loss of at least $\gamma$. Let a linear regressor $y = \br^{\sf T}\bx$ such that $\sum_{i=1}^n\left(y^{(i)} - \br^{\sf T}\bx^{(i)}\right)^2 = \omega n$. If $\ol{\br}$ is the closest to $\br$ in an $\nu \gamma/2B_{\sf l}B_{\sf X}$-net over $\mathbb{B}^d_{2}(B_{\sf r})$ then $(1-\nu)\omega n \leq \sum_{i=1}^n\left(y^{(i)} - \ol{\br}^{\sf T}\bx^{(i)}\right)^2 \leq (1+\nu)\omega n$.
\end{lemma}
\begin{proof} It is clear from the following
\begin{align}
    \left|\sum_{i=1}^n\left(y^{(i)} - \ol{\br}^{\sf T}\bx^{(i)}\right)^2 - \omega n\right| & = \left|\sum_{i=1}^n\left(y^{(i)} - \br^{\sf T}\bx^{(i)}\right)^2 - \sum_{i=1}^n\left(y^{(i)} - \ol{\br}^{\sf T}\bx^{(i)}\right)^2\right| \nonumber \\
    \leq 2B_{\sf l}\sum_{i=1}^n\left|(\br - \ol{\br})^{\sf T}\bx^{(i)}\right| & \leq 2B_{\sf l}\sum_{i=1}^n\|(\br - \ol{\br})\|_2\|\bx^{(i)}\|_2 \leq \nu\gamma n \leq \nu\omega n \nonumber
\end{align}
\end{proof}
\begin{lemma}\label{lem:bag_net_to_general_general}
Let a dataset $\{\bx^{(i)}, y^{(i)}\}_{i=1}^n$ and a linear regressor $y = \br^{\sf T}\bx$. Let $\{\bx_{\sf AGG}^{(j)}, y_{\sf AGG}^{(j)}\}_{j=1}^m$ be the dataset returned by \LBA with bag size $k$. Let $\mc{T}$ be a $(\nu\gamma/2m^2kB_{\sf l}B_{\sf X})$-net over $\mathbb{B}^{d}_{2}(B_{\sf r})$. Let $\ol{\br}$ be the closest to $\br$ in $\mc{T}$ such that $\sum_{j=1}^m\left(y_{\sf AGG}^{(j)} - \ol{\br}^{\sf T}\bx_{\sf AGG}^{(j)}\right)^2 = \omega mk$ where $\omega \geq \gamma$ for all $\ol{\br} \in \mc{T}$. Then w.p. at least $1 - k\cdot\sqrt{2/\pi}\exp(-m^2/2)$
\begin{equation}
    (1-\nu)\omega mk \leq \sum_{j=1}^m\left(y_{\sf AGG}^{(j)} - \br^{\sf T}\bx_{\sf AGG}^{(j)}\right)^2 \leq (1 + \nu)\omega mk\nonumber
\end{equation}
\end{lemma}
\begin{proof} Using the triangular inequality, 
\begin{align}
    \left|\sum_{j=1}^m\left(y_{\sf AGG}^{(j)} - \br^{\sf T}\bx_{\sf AGG}^{(j)}\right)^2 - \omega mk\right| = \left|\sum_{j=1}^m\left(y_{\sf AGG}^{(j)} - \br^{\sf T}\bx_{\sf AGG}^{(j)}\right)^2 - \sum_{j=1}^m\left(y_{\sf AGG}^{(j)} - \ol{\br}^{\sf T}\bx_{\sf AGG}^{(j)}\right)^2\right| \nonumber \\
    \leq \sum_{j=1}^m|2y_{\sf AGG}^{(j)} - (\br + \ol{\br})^{\sf T}\bx_{\sf AGG}^{(j)}|\|(\br - \ol{\br})\|_2\|\bx_{\sf AGG}^{(j)}\|_2 \label{eq:agg_loss_net_bound_general}
\end{align}
Using the concentration of gaussians, a standard gaussian $g$ satisfies the following
\begin{equation}
    \Pr\left[|g| \leq m\right] \geq 1 - \tfrac{1}{m}\sqrt{\tfrac{2}{\pi}}\exp(-m^2/2) \nonumber
\end{equation}
Taking a union bound over all gaussian weights $w_{jr}$ for all $j = 1, \dots, m$ and $r = 1, \dots, k$,
\begin{equation}\label{eq:agg_loss_net_union_general}
    \Pr[E_1] = \Pr\left[\wedge_{j=1}^m\wedge_{r=1}^k\left(|w_{jr}| \leq m\right)\right] \geq 1 - k\cdot\sqrt{\tfrac{2}{\pi}}\exp(-m^2/2) \nonumber
\end{equation}
If $E_1$ is satisfied then for all $j = 1, \dots, m$ then $\|\bx_{\sf AGG}^{(j)}\|_2 \leq mkB_2$ and $|y_{\sf AGG}^{(j)} - \br^{\sf T}\bx_{\sf AGG}^{(j)}| \leq 2mkB_{\sf l}$. Thus by \eqref{eq:agg_loss_net_bound_general}, 
\begin{equation}
    \left|\sum_{j=1}^m\left(y_{\sf AGG}^{(j)} - \br^{\sf T}\bx_{\sf AGG}^{(j)}\right)^2 - \omega mk\right| \leq m(2mkB_{\sf l})(\nu\alpha/2m^2kB_{\sf l}B_{\sf X})(mkB_{\sf X}) \leq \nu\alpha mk \leq \nu \omega mk \nonumber
\end{equation}
\end{proof}

\begin{theorem}
Let a dataset $\{\bx^{(i)}, y^{(i)}\}_{i=1}^n$ such that $\sum_{i=1}^n\left(y^{(i)} - \br^{\sf T}\bx^{(i)}\right)^2 \geq \gamma n$ for all linear regressors $y = \br^{\sf T}\bx$. Let $\{\bx_{\sf AGG}^{(j)}, y_{\sf AGG}^{(j)}\}_{j=1}^m$ be the dataset returned by \LBA with bag size $k$. Then for all linear regressors $y = \br^{\sf T}\bx$ with $\|\br\|_2 \leq B_{\sf r}$,
\begin{equation}
    \tfrac{1}{mk}\sum_{j=1}^m\left(y_{\sf AGG}^{(j)} - \br^{\sf T}\bx_{\sf AGG}^{(j)}\right)^2 \in \left(\tfrac{1}{n}\sum_{i=1}^n\left(y^{(i)} - \br^{\sf T}\bx^{(i)}\right)^2\right)\cdot(1 - \theta, 1 + \theta) \nonumber
\end{equation}
except with probability
\begin{equation}
    \exp\left(-\min\left\{\Omega\left(\tfrac{mk\theta^2\gamma^2}{B_{\sf l}^4}\right) - O\left(d\log\left(\tfrac{m^2kB_{\sf l}^2}{\theta\gamma}\right)\right), \Omega(m^2) - O(\log k)\right\}\right) \nonumber
\end{equation}
\end{theorem}
\begin{proof}
Let $\mc{T}$ be an $(\nu\gamma/4m^2kB_{\sf l}B_{\sf X})$-net over $\mathbb{B}^d_{2}(B_{\sf r})$. Using Lemma \ref{lem:covering_no_ball} and using the fact that $B_{\sf r}B_{\sf X} \leq B_{\sf l}$, $|\mc{T}| \leq O\left(m^2kB_{\sf l}^2/\nu\gamma\right)^d$. Using Lemma \ref{lem:single_regressor_bound_general} and taking a union bound over all linear regressors with $\ol{\br} \in \mc{T}$,
\begin{equation}
    \tfrac{1}{mk}\sum_{j=1}^m\left(y_{\sf AGG}^{(j)} - \ol{\br}^{\sf T}\bx_{\sf AGG}^{(j)}\right)^2 \in \left(\tfrac{1}{n}\sum_{i=1}^n\left(y^{(i)} - \ol{\br}^{\sf T}\bx^{(i)}\right)^2\right)\cdot(1 - \theta/6, 1 + \theta/6) \label{eq:utility_bound_1_general}
\end{equation}
for all $\ol{\br} \in \mc{T}$ except with probability at most $\exp\left(-\Omega\left(\tfrac{m\theta^2\gamma^2}{B_{\sf l}^4}\right) + O\left(d \log\left(\tfrac{m^2kB_{\sf l}^2}{\nu\gamma}\right)\right)\right)$. Now let $\br \in \mathbb{B}^d_{2}(B_{\sf r})$. Let $\ol{\br}$ be the closest vector to $\br$ in $\mc{T}$. Since any linear regressor has an average loss of at least $\gamma$, using Lemma \ref{lem:instance_general_to_net_general}, 
\begin{equation}
    \tfrac{1}{n}\sum_{i=1}^n\left(y^{(i)} - \ol{\br}^{\sf T}\bx^{(i)}\right)^2 \in \left(\tfrac{1}{n}\sum_{i=1}^n\left(y^{(i)} - \br^{\sf T}\bx^{(i)}\right)^2\right)\cdot\left(1 - \tfrac{\nu}{2m^2k}, 1 + \tfrac{\nu}{2m^2k}\right) \label{eq:utility_bound_2_general}
\end{equation}
If the event in \eqref{eq:utility_bound_1_general} occurs with $\theta \leq 1/4$, then $\sum_{j=1}^m\left(y_{\sf AGG}^{(i)} - \ol{\br}^{\sf T}\bx_{\sf AGG}^{(j)}\right)^2 \geq \alpha mk/2$ for all $\ol{\br} \in \mc{T}$. Then, using Lemma \ref{lem:bag_net_to_general_general}, except with probability at most $k\cdot\sqrt{2/\pi}\exp(-m^2/2)$.
\begin{equation}
    \tfrac{1}{mk}\sum_{j=1}^m\left(y_{\sf AGG}^{(j)} - \br^{\sf T}\bx_{\sf AGG}^{(j)}\right)^2 \in \left(\tfrac{1}{mk}\sum_{j=1}^m\left(y_{\sf AGG}^{(j)} - \br^{\sf T}\bx_{\sf AGG}^{(j)}\right)^2\right)\cdot(1 - \nu, 1 + \nu) \label{eq:utility_bound_3_general}
\end{equation}
Taking a union bound over events in \eqref{eq:utility_bound_1_general}, \eqref{eq:utility_bound_2_general}, \eqref{eq:utility_bound_3_general}, except with probability
\begin{equation}
    \exp\left(-\min\left\{\Omega\left(\tfrac{mk\theta^2\gamma^2}{B_{\sf l}^4}\right) + O\left(d \log\left(\tfrac{m^2kB_{\sf l}^2}{\nu\gamma}\right)\right), \Omega(m^2) - O(\log k)\right\}\right) \nonumber
\end{equation}
for all $\br \in \mathbb{B}^d_{2}(B_3)$
\begin{equation}
    \tfrac{1}{mk}\sum_{j=1}^m\left(y_{\sf AGG}^{(j)} - \br^{\sf T}\bx_{\sf AGG}^{(j)}\right)^2 \in \left(\tfrac{1}{n}\sum_{i=1}^n\left(y^{(i)} - \br^{\sf T}\bx^{(i)}\right)^2\right)\cdot(a, b)
\end{equation}
where $a = (1-\nu/2m^2k)(1-\theta/6)(1-\nu)$ and $b = (1+\nu/2m^2k)(1+\theta/6)(1+\nu)$. Picking $\nu = \eps/6$, we get our result since $k, m \geq 1$.
\end{proof}

\section{Privacy Analysis of \LLP}\label{sec:privacy-noisy_wtd_llp}
\begin{theorem}
The output of \LLP on the domain set $[-B_{\sf Y}, B_{\sf Y}]^n$, given input parameters $m, k$ and any $\mc{N} \subseteq [n]$ such that $\rho = |\mc{N}|/n$, is $(\eps, \delta)$-label DP if $k \geq O(B_{\sf Y}^2/\rho\eps^2 + B_{\sf Y}^4/\rho^2\eps^4)$ and
\begin{equation}
    \delta = \exp\left(-\min\left\{\Omega\left(\tfrac{\eps^2\rho\sqrt{k}}{B_{\sf Y}^2}\right), \Omega(\rho^2k) - O(\log m), \Omega(\sqrt{k}) - O(\log m)\right\}\right) \nonumber
\end{equation}
\end{theorem}
\begin{proof}
Let $\mc{D}_{\by, S_j, \bw_j}^{(j)}$ be the distribution of $y_{\sf AGG}^{(j)}$ given the bag $S_j$ and the aggregation coefficients $\bw_j$. WLOG assume that for every $S_j$, there is an $r_{0j} \in \{0, \dots, k\}$ such that $\{i_{j1}, \dots, i_{jr_{0j}}\} \subseteq \mc{N}$. Thus, $\mc{D}_{\by, S_j, \bw_j}^{(j)}$ is a gaussian distribution with mean $\sum_{r=1}^k w_{jr}y^{(i_{jr})}$ and variance $\sum_{r = 1}^{r_{0j}} |w_{jr}|^2$. Let $\mc{D}_{\by, S_j, \bw_j}^{(j)} \sim N(\mu_1, \sigma_1^2)$ and $\mc{D}_{\ol{\by}, S_j, \bw_j}^{(j)} \sim N(\mu_2, \sigma_2^2)$ where $\|\by - \ol{\by}\|_0 \leq 1 \Rightarrow\|\by - \ol{\by}\|_2 \leq 2B_1$.
\begin{equation}\label{eq:noisy_llp_conditional_tv_bound}
    |\mu_1 - \mu_2| \leq 2B_1\left(\underset{r = 1, \dots, k}{\max}\left\{|w_{jr}|\right\}\right) \qquad \sigma_1 = \sigma_0 = \sqrt{\sum_{i = 1}^{r_{0j}}|w_{jr}|^2} \nonumber
\end{equation}
Using concentration of gaussian, for any $j \in \{1, \dots, m\}$ and $r \in \{1, \dots, k\}$, 
\begin{equation}
    \Pr[|w_{jr}| \leq k^{\frac{1}{4}}] \geq 1 - \sqrt{2/\pi}\cdot\exp(-\sqrt{k}/2) \nonumber
\end{equation}
Take a union bound over all $j \in \{1, \dots, m\}$ and $r \in \{1, \dots, k\}$ and let that event be $E_1$. Then,
\begin{equation}
    \Pr[E_1] = \Pr\left[\wedge_{j=1}^m\wedge_{r=1}^k\left(|w_{jr}| \leq k^{\frac{1}{4}}\right)\right] \geq 1 - mk\sqrt{2/\pi}\exp(-\sqrt{k}/2) \nonumber
\end{equation}
Define a random variable $X_i$ which takes value $1$ if $i \in \mc{N}$ and $0$ otherwise. Thus, $\Ex[X_i] = \rho$. Using Theorem \ref{thm:Hoeffing} and in that using $\eps = \rho/2$, at least $\rho/2$ fraction of the points in $S_j$ are in $\mc{N}$ w.p. at least $1 - \exp(-k\rho^2/2)$. Taking a union bound over all bags, $r_{0j} \geq \rho k/2$ for all $j \in \{1, \dots, m\}$ w.p. at least $1 - m\cdot\exp(-k\rho^2/2)$. Let this event be $E_2$.\\
Define a random variable $G_j = \sum_{i = 1}^{r_{0j}}|w_{jr}|^2$. Thus, $\E_{w_{jr}}[G_j] = r_{0j}$. We can apply theorem \ref{lem:Gaussian-conc} since $G_j$ is the sum of squares of independent gaussians. Taking $\bv = (1)_{i=1}^{r_{0j}}$, we get
\begin{equation}
    \Pr\left[G_j \leq (1-\theta)r_{0j}\right] \leq \exp\left(-\tfrac{1}{2}\min\left\{\tfrac{\theta^2r_{0j}^2}{r_{0j}}, \tfrac{\theta r_{0j}}{1}\right\}\right) = \exp\left(-\tfrac{\theta^2r_{0j}}{2}\right) \nonumber
\end{equation}
Take a union bound of $|G_j \geq r_{0j}/2|$ over all $j \in \{1, \dots, m\}$ and $E_2$ and call this event $E_3$. Then,
\begin{equation}
    \Pr[E_3] = \Pr\left[\wedge_{j=1}^m\left(G_j \geq \rho k/4\right)\right] \geq 1 - m\cdot\exp\left(-\rho k/16\right) - m\cdot\exp\left(-k\rho^2/2\right) \nonumber
\end{equation}
When $E_1$ and $E_3$ both occur, for all $j \in \{1, \dots, m\}$, we get that
\begin{equation}
    \tfrac{|\mu_1 - \mu_2|}{\sigma_2} \leq 2B_{\sf Y}k^{\frac{1}{4}}/\sqrt{\gamma k / 4} = 4B_{\sf Y}/\gamma^{\frac{1}{2}}k^{\frac{1}{4}}  \nonumber
\end{equation}
In Lemma \ref{lem:Gaussian-deviation-single-dim}, pick $\kappa = 8B_1/\sqrt{\rho k}$, $\beta = \rho^{\frac{1}{4}}/\sqrt{2B_1}$ and $\theta = O(\eps\beta)$. All the conditions are satisfied when $k \geq O(B_{\sf Y}^2/\rho\eps^2 + B_{\sf Y}^4/\rho^2\eps^4)$. Thus, for any $S \subseteq [-B_{\sf Y}, B_{\sf Y}]$, $\Pr_{\mc{D}_{\by, S_j, \bw_j}}[S] \leq \exp(\eps)\Pr_{\mc{D}_{\ol{\by}, S_j, \bw_j}}[S] + \delta_1$ where $\delta_1 = \exp(-\Omega(\eps^2\rho\sqrt{k}/B_1^2))$ if $E_1$ and $E_3$ both occur.\\
Let $\mc{D}_{\by}^{(j)}$ be the distribution of $y_{\sf AGG}^{(j)}$ over $[-B_{\sf Y}, B_{\sf Y}]$ and $\hat{\mc{D}}$ be the distribution over $[n]^k \times \R^k$ of sampling of bags and weights. Let any subset $T^{(j)} \subseteq [-B_{\sf Y}, B_{\sf Y}] \times [n]^k \times \R^k$ and $T^{(j)}_{S_j, \bw_j} = \{y \,\mid\, (y, S_j, \bw_j) \in T^{(j)}\}$. Then,
\begin{align}
    \Pr_{\mc{D}_{\by}^{(j)}}[T^{(j)}] & = \E_{(S_j, \bw_j) \leftarrow \hat{\mc{D}}}\left[\Pr_{\mc{D}_{\by, S_j, \bw_j}^{(j)}}[T^{(j)}_{S_j, \bw_j}]\right] \nonumber \\
    & \leq \E_{(S_j, \bw_j) \leftarrow \hat{\mc{D}}}\left[\Pr_{\mc{D}_{\by, S_j, \bw_j}^{(j)}}[T^{(j)}_{S_j, \bw_j} \mid \mathbbm{1}_{E_1, E_3}] + \mathbbm{1}_{E_1, E_3}\right] \nonumber \\
    & \leq \E_{(S_j, \bw_j) \leftarrow \hat{\mc{D}}}\left[e^{\eps}\Pr_{\mc{D}_{\ol{\by}, S_j, \bw_j}^{(j)}}[T^{(j)}_{S_j, \bw_j} \mid \mathbbm{1}_{E_1, E_3}] + \delta_1\right] + \E_{(S_j, \bw_j) \leftarrow \hat{\mc{D}}}\left[\mathbbm{1}_{E_1, E_3}\right] \nonumber \\
    & \leq \exp(\eps)\Pr_{\mc{D}_{\ol{\by}}^{(j)}}[T^{(j)}] + \delta_1 + m\exp(-\rho k/16) + m\eps(-\rho k^2/2) + mk\exp(-\sqrt{k}/2) \nonumber
\end{align}
Thus, $\Pr_{\mc{D}_{\by}^{(j)}}[T^{(j)}] \leq \exp(\eps)\Pr_{\mc{D}_{\ol{\by}}^{(j)}}[T^{(j)}] + \delta$ where
\begin{equation}
    \delta = \exp\left(-\min\left\{\Omega\left(\tfrac{\eps^2\rho\sqrt{k}}{B_1^2}\right), \Omega(\rho^2k) - O(\log m), \Omega(\sqrt{k}) - O(\log m)\right\}\right) \nonumber
\end{equation}
Now, since $S_1, \dots, S_m$ are disjoint the label DP guarantee follows from the single bag bound above as a change in a single $y^{(i)} (i \in [n])$ affects at most one bag.
\end{proof}

\section{Neural Network Setup}\label{sec:neural-net-noisy_wtd_llp}

We will show utility bound over a general neural network. The dataset setup is the same as that in Sec. \ref{sec:Ourresults} for \LLP. We assume that for a given dataset $\mb{D} := \{\bx^{(i)}, y^{(i)}\}_{i=1}^m$, $y^{(i)} \in [-B_{\sf Y}, B_{\sf Y}]$ and $\|\bx^{(i)}\|_2 \leq B_{\sf X}$ for all $i = 1, \dots, n$. We consider the class $\mc{F}_N$ of neural-networks $f$  characterized by its $d'$ weights $\mb{s}_f$ which satisfy: $\|\mb{s}_f\|_2 \leq B_4$ for all $f \in \mc{F}_N$, and the following Lipshcitzness property: $\left|f(\bx) - f'(\bx)\right| \leq L\|\mb{s}_f - \mb{s}_{f'}\|_2$, for $f, f' \in \mc{F}_1$. We use $B_{\tn{loss}}$ as an upper bound on squared loss on any point of $\mb{D}$. We will prove these properties for bounded neural networks. The output of the algorithm is an LLP dataset. Given access to it, the objective is to minimize $\val_{\sf LLP}(\tilde{\mb{D}}_{\tn{LLP}}, f) = \sum_{j=1}^m(\ol{y}^{(j)} - \sum_{i \in S_j}w_{ij}f(\bx^{(i)}))^2$, where $\opt_{\sf LLP}(\tilde{\mb{D}}_{\tn{LLP}}, \mc{F}_1) := \inf_{f \in \mc{F}_1}\val_{\sf LLP}(\tilde{\mb{D}}_{\tn{LLP}}, f)$.

\subsection{Properties of bounded neural networks}\label{subsec:llp_utlilty_prelim}
Define a function $f_l$ as follows
\begin{equation}
    f_l(\bx; \bW) = 
    \begin{cases}
        \bW_l[o(f_{l-1}(\bx, \bW))\,\mid\,1] & \text{ if } l > 0\\
        \bx & \text{ if } l = 0
    \end{cases} \nonumber
\end{equation}
A neural network $f$ of $l_0$ layers can be defined as $f = f_{l_0}$. Here $[ \cdot \mid 1]$ represents the operation of appending $1$ to a vector, $o(\cdot)$ is the activation function which is $1$-lipschitz and $\bW_l$ is the weight matrix of the $l^{th}$ layer. We say this neural network is bounded of the frobenius norm of every $\bW_l$ is bounded by some positive constant $K$. Let us call this class of neural networks $\mc{F}_{\tn{NN}}(K, l_0)$. WLOG let $K \geq B_2$ (if that is not the case then take $K = \max(K, B_2)$. Observe that if $f \in \mc{F}_{\tn{NN}}(K, l_0)$ then, $\|\bs_f\|_2 \leq \sqrt{l_0}K$.

\begin{lemma}
The output of neural networks in $\mc{F}_{\tn{NN}}(K, l_0)$ is bounded by $K\sqrt{\tfrac{K^{2l_0+2} - 1}{K^2-1}}$. It follows that the mean squared error is bounded by $\left(B_1 + K\sqrt{\tfrac{K^{2l_0+2} - 1}{K^2-1}}\right)^2$.
\end{lemma}
\begin{proof}
Using Cautchy-Schwartz inequality and the fact that $\|o(\bx)\|_2 \leq \|\bx\|_2$, for all $l > 0$,
\begin{align}
    \|[f_l(\bx; \bW)\|_2^2 & \leq \|\bW_l\|_2^2(\|o(f_{l-1}(\bx; \bW))\|_2^2+1) \nonumber \\
    & \leq \|\bW_l\|_F^2(\|f_{l-1}(\bx; \bW)\|_2^2+1) \nonumber \\
    &  \leq K^2(\|f_{l-1}(\bx; \bW)\|_2^2+1) \nonumber
\end{align}
and $\|f_0(\bx; \bW)\|_2^2 = \|\bx\|_2^2 \leq K^2$. Unrolling the recursion, $\|[f_{l_0}(\bx; \bW)\|_2^2 \leq K^2\tfrac{K^{2l_0 + 2} - 1}{K^2 - 1}$.
\end{proof}

\begin{lemma}
\label{lem:lipschitzness_neural_net}
$f \in \mc{F}_{\tn{NN}}(K, l_0)$ are $\max\left\{\sqrt{l_0\tfrac{K^{2l+2}-1}{K^2-1}}, l_0\right\}$-lipschitz w.r.t the weights.
\end{lemma}
\begin{proof}
For convenience, let $f = f_{l_0}(\cdot, \bW)$ and $f' = f_{l_0}(\cdot, \bW')$. Using the triangular inequality, $1$-lipschitzness of $o(\cdot)$ for all $l = \{1, \dots, l_0\}$
\begin{align}
    \|f_l(\bx) - f_l'(\bx)\|_2 & \leq \|\bW_l\|_2\|f_{l-1}(\bx) - f'_{l-1}(\bx)\|_2 + \|\bW_l - \bW'_l\|_2\|[f'_{l-1}(\bx)|1]\|_2 \nonumber \\
    & \leq K\|f_{l-1}(\bx) - f'_{l-1}(\bx)\|_2 + \|\bW_l - \bW'_l\|_2\sqrt{\tfrac{K^{2l+2}-1}{K^2-1}} \nonumber
\end{align}
and $\|f_0(\bx) - f_0'(\bx)\|_2 = 0$. Unrolling the recursion, we get that
\begin{align}
    \|f_{l_0}(\bx) - f_{l_0}'(\bx)\|_2 & \leq 
    \begin{cases}
    \sqrt{\tfrac{K^{2l+2}-1}{K^2-1}}\left(\sum_{l=1}^{l_0}\|\bW_l - \bW'_l\|_2\right) & \text{ if } K > 1\\
    \sum_{l=1}^{l_0}\|\bW_l - \bW'_l\|_2 & \text{ if } K \leq 1\\
    \end{cases} \nonumber \\
     & \leq \begin{cases}
    \sqrt{\tfrac{K^{2l+2}-1}{K^2-1}}\left(\sum_{l=1}^{l_0}\|\bW_l - \bW'_l\|_F\right) & \text{ if } K > 1\\
    \sqrt{l_0}\sum_{l=1}^{l_0}\|\bW_l - \bW'_l\|_F & \text{ if } K \leq 1\\
    \end{cases} \nonumber \\
    & \leq \begin{cases}
    \sqrt{l_0\tfrac{K^{2l+2}-1}{K^2-1}}\|\bs_f - \bs_f'\|_2 & \text{ if } K > 1\\
    l_0\|\bs_f - \bs_f'\|_2 & \text{ if } K \leq 1\\
    \end{cases} \nonumber
\end{align}
\end{proof}

\section{Utility Analysis of \LLP}\label{sec:utility-noisy_wtd_llp}
In this section, we shall prove that $\tfrac{1}{n}\val(\tilde{\mb{D}}, f) - \theta \leq \tfrac{1}{mk}\val_{\sf LLP}(\tilde{\mb{D}}_{\sf LLP}, f) \leq \tfrac{1}{n}\val(\tilde{\mb{D}}, f) + \theta$ for all $f \in \mc{F}_N$ with probability at least that mentioned in Theorem \ref{thm-intro-main3}. The rest of the proof follows from the following argument. Let $f^* = {\sf argmin}_{f \in \mc{F}_N}\val_{\sf LLP}(\tilde{\mb{D}}_{\sf LLP}, f)$ and $f^{**} = {\sf argmin}_{f \in \mc{F}_N}\val(\tilde{\mb{D}}, f)$, then our result follows by the following argument.
\begin{align}
    \tfrac{1}{n}\val(\tilde{\mb{D}}, f^*) - \theta \leq \tfrac{1}{mk}\val_{\sf LLP}(\tilde{\mb{D}}_{\sf LLP}, f^*) \leq \tfrac{1}{mk}\val_{\sf LLP}(\tilde{\mb{D}}_{\sf LLP}, f^{**}) \leq \tfrac{1}{n}\val(\tilde{\mb{D}}, f^{**}) + \theta \nonumber
\end{align}

\begin{lemma}\label{lem:instance_level_utility_nn}
Given a dataset $\mb{D} = \{\bx^{(i)}, y^{(i)}\}_{i=1}^m$ and a neural network $f \in \mc{F}_N$. Let $\tilde{\mb{D}} = \{\bx^{(i)}, \tilde{y}^{(i)}\}$ be the intermediate dataset generated by \LLP with any arbitary $\mc{N}$ such that $|\mc{N}|/n = \rho$. Then $|\tfrac{1}{n}\val(\tilde{\mb{D}}, f) - (\rho + \tfrac{1}{n}\val(\mb{D}, f))| \leq \theta$ except with probability $\exp(-\Omega(\theta^2 n/\rho B_{\tn{loss}}))$.
\end{lemma}
\begin{proof}
Observe that 
\begin{equation}
|\tfrac{1}{n}\val(\tilde{\mb{D}}, f) - (\rho + \tfrac{1}{n}\val(\mb{D}, f))| \leq |\tfrac{1}{n}\sum_{i \in \mc{N}}g_i^2 - \rho| + |\tfrac{1}{n}\sum_{i \in \mc{N}}2g_i(y^{(i)} - f(\bx^{(i)}))|    \nonumber
\end{equation}
The first term is the sum of squares of gaussians. Using Lemma \ref{lem:Gaussian-conc} with $\bv = (1)_{1}^{\rho n}$,
\begin{equation}
    \Pr\left[\left|\sum_{i \in \mc{N}}g_i^2 - \rho n\right| \geq \tfrac{\theta n}{2}\right] \leq \exp\left(-\tfrac{1}{2}\min\left\{\tfrac{\theta^2n^2}{4\rho n}, \tfrac{\theta n}{2}\right\}\right) \leq \exp\left(-\tfrac{\theta^2n}{8\rho}\right) \nonumber
\end{equation}
The second term is a gaussian with variance at most $4B_{\tn{loss}}\rho/n$. Hence,
\begin{equation}
    \Pr\left[\left|\tfrac{1}{n}\sum_{i \in \mc{N}}2g_i(y^{(i)} - f(\bx^{(i)}))\right| \geq \tfrac{\theta}{2}\right] \leq \exp\left(-\tfrac{\theta^2n}{8\rho B_{\tn{loss}}}\right) \nonumber
\end{equation}
This concludes the proof.
\end{proof}

\begin{lemma}\label{lem:bag_level_utility_nn}
Given a dataset $\mb{D} = \{\bx^{(i)}, y^{(i)}\}_{i=1}^m$ and a neural network $f \in \mc{F}_N$. Let $\tilde{\mb{D}}_{\tn{LLP}}$ be the output of \LLP with $m, k, \mc{N}$ such that $|\mc{N}|/n = \rho$. Then, $\left|\tfrac{1}{mk}\val_{\sf LLP}(\tilde{\mb{D}}_{\tn{LLP}}, f) - (\rho + \tfrac{1}{n}\val(\mb{D}, f))\right| \leq \theta$ except with probability
\begin{equation}
    \exp\left(- \min \left\{\Omega\left(\tfrac{m\theta^2}{B_{\tn{loss}}^2}\right), \Omega(m^{1/4}) - O(\log k), \Omega\left(\tfrac{\theta^2\sqrt{m}}{k^2 + kB_{\tn{loss}}}\right)\right\}\right) \nonumber
\end{equation}
\end{lemma}
\begin{proof} Notice that the term to be bounded is lower bounded by the sum of the following three terms. The first term is
\begin{equation}
     T_1 = \left|\tfrac{1}{mk}\sum_{j=1}^m\left(\sum_{r=1}^kw_{jr}\left(y^{(i_{jr})} - f(\bx^{(i_{jr})}) \right)\right)^2 - \tfrac{1}{n}\sum_{i=1}^n\left(y^{(i)} - f(\bx^{(i)})\right)^2\right| \nonumber
\end{equation}
Define $\omega = \tfrac{1}{n}\sum_{i=1}^n\left(y^{(i)} - f(\bx^{(i)})\right)^2$ and $G_j = \sum_{r=1}^kw_{jr}\left(y^{(i_{jr}) - f(\bx^{(i_{jr})})} \right)$. Notice that $\E_{w_{jr}}\left[\tfrac{1}{mk}\sum_{j=1}^mG_j^2\right] = \tfrac{1}{mk}\sum_{j=1}^m\sum_{r=1}^k\left(y^{(i_{jr})} - f(\bx^{(i_{jr})})\right)^2$. Since $0 \leq \left(y^{(i_{jr})} - f(\bx^{(i_{jr})})\right)^2 \leq B_{\tn{loss}}$, we can use Theorem \ref{thm:Hoeffing} to obtain that
\begin{equation}
    \Pr\left[\left|\E_{w_{jr}}\left[\tfrac{1}{mk}\sum_{j=1}^mG_j^2\right] - \omega\right| \geq \theta\right] \leq \exp\left(-\tfrac{2m^2k^2\theta^2}{mkB_{\tn{loss}}}\right) = \exp\left(-\tfrac{2mk\theta^2}{B_{\tn{loss}}}\right) \nonumber
\end{equation}
Now, $G_j$ is a gaussian with variance $\sum_{i=1}^k(y^{(i_{jr})} - f(\bx^{(i_{jr})}))^2 \leq kB_{\tn{loss}}$. Hence, $\sum_{j=1}^mG_j^2$ is the sum of square of gaussians with $\bv = (\E[G_j^2])_{j=1}^m$. Thus, $\|\bv\|_2^2 \leq mk^2B_{\tn{loss}}^2$ and $\|\bv\|_\infty \leq kB_{\tn{loss}}$. Using Lemma \ref{lem:Gaussian-conc},
\begin{equation}
    \Pr\left[\left|\tfrac{1}{mk}\sum_{j=1}^mG_j^2 - \E_{w_{jr}}\left[\tfrac{1}{mk}\sum_{j=1}^mG_j^2\right]\right| \geq \theta\right] \leq \exp\left(-\tfrac{1}{2}\min\left\{\tfrac{m^2k^2\theta^2}{mk^2B_{\tn{loss}}^2}, \tfrac{mk\theta}{kB_{\tn{loss}}}\right\}\right) = \exp\left(-\tfrac{m\theta^2}{2B_{\tn{loss}}^2}\right) \nonumber
\end{equation}
Thus, $\Pr[T_1 \geq \theta] \leq \exp(-\Omega(m\theta^2/B_{\tn{loss}}^2))$. The second term is the absolute value of
\begin{equation}
    T_2 = \tfrac{2}{mk}\sum_{i=1}^m\sum_{r=1}^{r_{0j}}G_{jr} \qquad G_{jr} = w_{jr}\left(\sum_{r=1}^kw_{jr}(y^{(i_{jr})} - f(\bx^{(i_{jr})}))\right)g_{i_{jr}} \nonumber
\end{equation}
Notice here that $G_{jr} \sim N(0, w_{jr}^2(\sum_{r=1}^kw_{jr}(y^{(i_{jr})} - f(\bx^{(i_{jr})})))^2)$. If $|w_{jr}| \leq m^{1/8}$ then the variance is bounded by $\sqrt{m}k^2B_{\tn{loss}}$. This happens with probability at least $1 - mk\exp(\Omega(m^{1/4}))$. Then,
\begin{equation}
    \Pr[|T_2| \geq \theta] \leq \exp\left(-\Omega(\tfrac{m^2k^2\theta^2}{\sqrt{m}k^2B_{\tn{loss}}\sum_{j=1}^mr_{0j}})\right) \leq \exp\left(-\Omega(\tfrac{m^2\theta^2}{\sqrt{m}B_{\tn{loss}}mk})\right) = \exp\left(-\Omega(\tfrac{\sqrt{m}\theta^2}{kB_{\tn{loss}}})\right) \nonumber
\end{equation}
Taking a union bound, we get that $\Pr[|T_2| \geq \theta] \leq mk\exp(-\Omega(m^{1/4})) + \exp(-\Omega(\sqrt{m}\theta^2/kB_{\tn{loss}}))$. Finally the third term is of the form
\begin{equation}
    |T_3 - \rho| \qquad \text{where} \qquad T_3 = \tfrac{1}{mk}\sum_{j=1}^m\left(\sum_{r=1}^{r_{0j}}w_{jr}g_{i_{jr}}\right)^2 \nonumber
\end{equation}
Define $T_3^{(1)} = \E_{g}[T_3], T_3^{(2)} = \E_{w_{jr}}[T_3^{(1)}]$. Observe that $\E_{r_{j0}}[T_3^{(2)}] = \rho$. Using Theorem \ref{thm:Hoeffing}, we get
\begin{equation}
    \Pr[|T_3^{(2)} - \rho| \geq \theta] \leq \exp(-\Omega(mk\theta^2)) \nonumber
\end{equation}
$T_3^{(1)} = \tfrac{1}{mk}\sum_{j=1}^m\sum_{r=1}^{r_{0j}}w_{jr}^2$. Since this is the sum of square of gaussians, using Lemma \ref{lem:Gaussian-conc},
\begin{equation}
    \Pr[|T_3^{(1)} - T_3^{(2)}| \geq \theta] \leq \exp\left(-\tfrac{1}{2}\min\left\{\tfrac{m^2k^2\theta^2}{\sum_{j=1}^mr_{0j}}, mk\theta\right\}\right) \leq \exp\left(-\Omega(mk\theta^2)\right) \nonumber
\end{equation}
Finally, $\Pr[\wedge_{j=1}^m\wedge_{r=1}^k\left(|w_{tr}| \geq m^{1/8}\right)] \leq mk\exp(\Omega(m^{1/4}))$. Given this, $T_3$ is the sum of square of $m$ independent gaussians with $\bv = (\sum_{r=1}^{r_{0j}}w_{jr}^2)_{j=1}^m$. Thus, $\|\bv\|_2^2 \leq k^2m^{3/2}$ and $\|\bv\|_\infty \leq km^{1/4}$. Using Lemma \ref{lem:Gaussian-conc}, \begin{equation}
    \Pr[|T_3 - T_3^{(1)}| \geq \theta] \leq \exp\left(-\tfrac{1}{2}\min\left\{\tfrac{m^2\theta^2}{k^2m^{3/2}}, \tfrac{m\theta}{km^{1/4}}\right\}\right) \leq \exp(-\Omega(\sqrt{m}\theta^2/k^2)) \nonumber
\end{equation}
Taking a union bound, $\Pr[|T_3 - \rho| \geq \theta] \leq \exp(-\Omega(mk\theta^2)) + \exp(-\Omega(\sqrt{m}\theta^2/k^2))$. Taking a union bound over all three terms, we get that $\left|\tfrac{1}{mk}\val(\tilde{\mb{D}}_{\tn{LLP}}, f) - (\rho + \tfrac{1}{n}\val(\mb{D}, f))\right| \leq \theta$ except with probability
\begin{equation}
    \exp\left(- \min \left\{\Omega\left(\tfrac{m\theta^2}{B_{\tn{loss}}^2}\right), \Omega(m^{1/4}) - O(\log k), \Omega\left(\tfrac{\theta^2\sqrt{m}}{k^2 + kB_{\tn{loss}}}\right)\right\}\right) \nonumber
\end{equation}
\end{proof}

\begin{lemma}\label{lem:instance_to_instance_net_nn}
Let $\tilde{\mb{D}} = \{\bx^{(i)}, \tilde{y}^{(i)}\}_{i=1}^m$ be an intermediate dataset of \LLP on $\mb{D}$ with $\mc{N}$ as parameter such that $|\mc{N}|/n = \rho$. If $f'$ is a neural network with parameter closest to the parameters of $f$ in $\theta/2L\sqrt{B_{\tn{loss}}}$-net then $\left|\tfrac{1}{n}\val(\tilde{\mb{D}}, f) - \tfrac{1}{n}\val(\tilde{\mb{D}}, f')\right| \leq \theta$.
\end{lemma}
\begin{proof}
\begin{equation}
    \left|\tfrac{1}{n}\val(\tilde{\mb{D}}, f) - \tfrac{1}{n}\val(\tilde{\mb{D}}, f')\right| \leq \tfrac{1}{n}\sum_{i=1}^n(2\sqrt{B_{\tn{loss}}})|f - f'|(\bx^{(i)}) \leq \theta \nonumber
\end{equation}
\end{proof}

\begin{lemma}\label{lem:bag_to_bag_net_nn}
Let $\tilde{\mb{D}}_{\tn{LLP}}$ be the output of \LLP on $\mb{D}$ with $m, n, \mc{N}$ as parameter such that $|\mc{N}|/n = \rho$. If $f'$ is a neural network with parameter closest to the parameters of $f$ in $\theta/m^2(2\sqrt{B_{\tn{loss}}} + m)L$-net then $\left|\tfrac{1}{mk}\val_{\sf LLP}(\tilde{\mb{D}}_{\tn{LLP}}, f) - \tfrac{1}{mk}\val_{\sf LLP}(\tilde{\mb{D}}_{\tn{LLP}}, f')\right| \leq \theta$ w.p. at least $1 - 2mk\exp(-\Omega(m^2))$.
\end{lemma}
\begin{proof} $\Pr[\wedge_{j=1}^m\wedge_{r=1}^k((|g_{i_{jr}}| \geq m)\wedge(|w_{jr}| \geq m))] \leq 2mk\exp(-\Omega(m^2))$
\begin{align}
    \left|\tfrac{1}{mk}\val_{\sf LLP}(\tilde{\mb{D}}_{\tn{LLP}}, f) - \tfrac{1}{mk}\val_{\sf LLP}(\tilde{\mb{D}}_{\tn{LLP}}, f')\right| & \leq \tfrac{1}{mk}\sum_{j=1, r=1}^{m, k}w_{jr}^2(2\sqrt{B_{\tn{loss}}} + m)|f - f'|(\bx^{(i_{jr})}) \nonumber \\
    & \leq \tfrac{1}{mk}(mk)(m^2)(2\sqrt{B_{\tn{loss}}} + m)L\tfrac{\theta}{m^2(2\sqrt{B_{\tn{loss}}} + m)L} \nonumber \\
    & \leq \theta \nonumber
\end{align}
\end{proof}

\begin{lemma}
Let $\tilde{\mb{D}} = \{\bx^{(i)}, \tilde{y}^{(i)}\}_{i=1}^m$ be an intermediate dataset and $\tilde{\mb{D}}_{\tn{LLP}}$ be the output of \LLP on $\mb{D}$ with $m, n, \mc{N}$ as parameter such that $|\mc{N}|/n = \rho$. Then $\left|\tfrac{1}{mk}\val_{\sf LLP}(\tilde{\mb{D}}_{\tn{LLP}}, f) - \tfrac{1}{n}\val(\tilde{\mb{D}}, f)\right| \leq \theta$ for all $f \in \mc{F}_N$ except with probability at most
\begin{equation}
    \exp\left(-\min\left\{\Omega(\tfrac{m\theta^2}{B_{\tn{loss}}^2}), \Omega(\tfrac{ mk\theta^2}{\rho B_{\tn{loss}}}), \Omega(m^{1/4}) - O(\log k), \Omega(\tfrac{\theta^2\sqrt{m}}{k^2 + kB_{\tn{loss}}})\right\} + O\left(d' \log\left(\tfrac{mLB_4B_{\tn{loss}}}{\theta}\right)\right)\right) \nonumber
\end{equation}
\end{lemma}
\begin{proof}
Let $\mc{T}$ be a $\theta/m^2(2\sqrt{B_{\tn{loss}}} + m)L$-net over $B_{\ell_2}(0, B_s)$. Let $f'$ be a neural network with parameter in $\mc{T}$. Using Lemma \ref{lem:instance_level_utility_nn} and \ref{lem:bag_level_utility_nn}, $\left|\tfrac{1}{mk}\val_{\sf LLP}(\tilde{\mb{D}}_{\tn{LLP}}, f') - \tfrac{1}{n}\val(\tilde{\mb{D}}, f')\right| \leq \theta$ except w.p.
\begin{equation}
    \exp\left(-\min\left\{\Omega(\tfrac{m\theta^2}{B_{\tn{loss}}^2}), \Omega(\tfrac{ mk\theta^2}{\rho B_{\tn{loss}}}), \Omega(m^{1/4}) - O(\log k), \Omega(\tfrac{\theta^2\sqrt{m}}{k^2 + kB_{\tn{loss}}})\right\}\right) \nonumber
\end{equation}
Taking a union bound over $\mc{T}$, for all $f' \in \mc{T}$, $\left|\tfrac{1}{mk}\val_{\sf LLP}(\tilde{\mb{D}}_{\tn{LLP}}, f') - \tfrac{1}{n}\val(\tilde{\mb{D}}, f')\right| \leq \theta$ except w.p.
\begin{equation}
    \exp\left(-\min\left\{\Omega(\tfrac{m\theta^2}{B_{\tn{loss}}^2}), \Omega(\tfrac{ mk\theta^2}{\rho B_{\tn{loss}}}), \Omega(m^{1/4}) - O(\log k), \Omega(\tfrac{\theta^2\sqrt{m}}{k^2 + kB_{\tn{loss}}})\right\} + O\left(d' \log\left(\tfrac{mLB_sB_{\tn{loss}}}{\theta}\right)\right)\right) \nonumber
\end{equation}
Let $f \in \mc{F}_N$ be a neural network with parameters closest to parameters of $f'$. Using Lemma \ref{lem:instance_to_instance_net_nn} $\left|\tfrac{1}{n}\val(\tilde{\mb{D}}, f) - \tfrac{1}{n}\val(\tilde{\mb{D}}, f')\right| \leq \theta$. Also using Lemma \ref{lem:bag_to_bag_net_nn}, $\left|\tfrac{1}{mk}\val_{\sf LLP}(\tilde{\mb{D}}_{\tn{LLP}}, f') - \tfrac{1}{mk}\val(\tilde{\mb{D}}_{\tn{LLP}}, f)\right| \leq \theta$ w.p. at least $1 - 2mk\exp(-\Omega(m^2))$. Taking a union bound over all these three events, we get that $\left|\tfrac{1}{mk}\val_{\sf LLP}(\tilde{\mb{D}}_{\tn{LLP}}, f') - \tfrac{1}{n}\val(\tilde{\mb{D}}, f')\right| \leq \theta$ except w.p.
\begin{equation}
    \exp\left(-\min\left\{\Omega(\tfrac{m\theta^2}{B_{\tn{loss}}^2}), \Omega(\tfrac{ mk\theta^2}{\rho B_{\tn{loss}}}), \Omega(m^{1/4}) - O(\log k), \Omega(\tfrac{\theta^2\sqrt{m}}{k^2 + kB_{\tn{loss}}})\right\} + O\left(d' \log\left(\tfrac{mLB_sB_{\tn{loss}}}{\theta}\right)\right)\right) \nonumber
\end{equation}
\end{proof}

\section{Details of Experiments and Results} \label{sec:additional_exp}
\urlstyle{sf}
We use the following two datasets previously used by \citep{ghazi2022regression} for label-DP in regression settings. We keep most of the preprocessing same as in \citep{ghazi2022regression}.

{\bf Criteo Sponsored Search Conversion Logs Dataset~\citep{tallis2018reacting}.} Each datapoint in the dataset\footnote{\url{https://ailab.criteo.com/criteo-sponsored-search-conversion-log-dataset/}} represents a click by a user with conversion value within the next 30 days. We use the following features.
\begin{itemize}
    \item \emph{Target} : SalesAmountInEuro
    \item \emph{Numerical Features} : Time\_delay\_for\_conversion, nb\_clicks\_1week, product\_price
    \item \emph{Categorical Features} : product\_age\_group, product\_category\_1, product\_category\_7, product\_country, product\_gender
\end{itemize}
We remove the datapoints in which the target is missing. This leaves us with 1,732,721 labeled datapoints. On the categorical features, we group all the infrequent categorical values (occurring at most 5 times) along with the missing value (which is -1). We apply multihot encoding on them, making the encoded dimension 68. For the numerical features, we replace all missing values by the mean of that feature. We then preprocess them along with the target as described in \citep{song2019autoint}. The final dimension of the input is 71.

\medskip

{\bf 1940 US Census Data~\citep{IPUMS-USA}.}
This dataset\footnote{\url{https://usa.ipums.org/usa/1940CensusDASTestData.shtml}} is made publicly available since 2012. We download the census data of the year 1940 and use the following features.
\begin{itemize}
    \item \emph{Target} : WKSWORK - Number of weeks the person worked in the previous year
    \item \emph{Numerical Features} : AGE - Age of the person
    \item \emph{Categorical Features}: SEX - gender, MARST - marital status, CHBORN - number of children born to a woman in that year, SCHOOL - school attendance, EMPSTAT - employment status, OCC - primary occupation, IND - type of industry in which the person works
\end{itemize}
We again remove the datapoints with null labels leaving us with 50,582,693 datapoints. As before, the categorical features are encoded as multihot in a 401 dimensional space. The input dimension is 402.
\subsection{Training Details}
We create a train-test split of $80\%/20\%$. We then use the train part to create our aggregated datasets for \LBA and \LLP choosing values for $m, k$ and $\rho$ as reported below. We then train a linear model on \LBA dataset and a neural model with 128, 64 and 1 neurons in successive layers on \LLP dataset. We train the \LBA dataset using $\val$ as the loss and the \LLP dataset using the $\val$ modified for LLP as the loss. We train for 200 epochs, using the Adam optimizer. We use Cosine Decay for 200 epochs with initial learning rate of 1e-3 and alpha of 1e-3. For the Criteo Sponsored Search data, we use a batch size of 1024 and for the US census data we use a batch size of 8192. We use an early stopping callback with a patience parameter of 3 epochs. We report the MSE numbers on the test part. This is done over 10 random seeds and the mean and standard deviation are reported.
\subsection{Results}
Table \ref{tab:lin_lba_error} reports test MSE with different bag sizes and number of bags for \LBA. Consistent with our theoretical results, we observe that with increasing number of bags, the test loss of the regressor trained on the bags is closer to the instance-trained regressors test MSE. 

We report primary results for \LLP with $\rho=0.1$ in Table \ref{tab:noisy_llp_error}. We also perform additional experiments taking $\rho = 0.0, 0.01, 0.5$ and $1.0$ for \LLP which are reported in Tables \ref{tab:noisy_llp_gamma_0.0}, \ref{tab:noisy_llp_gamma_0.01}, \ref{tab:noisy_llp_gamma_0.5} and \ref{tab:noisy_llp_gamma_1.0} respectively. One can notice that there is not much change in the instance level MSE possibly since the targets have large range (Criteo Sponsored Search target 'SalesAmountInEuro' has a range of $[0,253.78]$ and US Census target 'WKSWORK' has a range of $[0,51]$). 

\begin{table}[!htb]
    \caption{Test MSE on training Linear model on data created using \LBA with error bars}
    \begin{minipage}{.5\linewidth}
        \centering
\resizebox{0.95\linewidth}{!}{
\begin{tabular}{lllll}
\toprule
\multicolumn{5}{c}{Instance Level MSE = 196.35\tiny{$\pm$0.88}} \\
\hline
\multicolumn{1}{c|}{} & \multicolumn{4}{c}{\textbf{k}}                          \\
\multicolumn{1}{c|}{\textbf{m}}             & \textit{32} & \textit{64} & \textit{128} & \textit{256} \\ \toprule
\textit{512}  & 252.9\tiny{$\pm$21.3}  & 234.61\tiny{$\pm$3.14} & 226.41\tiny{$\pm$3.06}  & 219.23\tiny{$\pm$2.26}  \\
\textit{1024} & 230.82\tiny{$\pm$3.63} & 220.88\tiny{$\pm$2.72} & 212.93\tiny{$\pm$1.22}  & 208.64\tiny{$\pm$1.63}  \\
\textit{2048} & 216.55\tiny{$\pm$2.13} & 208.88\tiny{$\pm$1.11} & 204.99\tiny{$\pm$1.25}  & 202.95\tiny{$\pm$0.76}  \\
\textit{4096} & 206.18\tiny{$\pm$0.81} & 202.87\tiny{$\pm$0.85} & 200.95\tiny{$\pm$0.65}  & 200.11\tiny{$\pm$0.92} \\
\bottomrule
\label{tab:lin_lba_criteo}

\end{tabular}
}
\subcaption{Criteo Sponsored Search Data}

    \end{minipage}%
    \begin{minipage}{.5\linewidth}
        \centering
\resizebox{0.95\linewidth}{!}{
\begin{tabular}{lllll} \toprule
\multicolumn{5}{c}{Instance Level MSE = 142.53\tiny{$\pm$0.08}} \\
\hline
\multicolumn{1}{c|}{} & \multicolumn{4}{c}{\textbf{k}}                          \\
\multicolumn{1}{c|}{\textbf{m}}             & \textit{64}  & \textit{128} & \textit{256} & \textit{512} \\ \toprule
\textit{1024} & 471.83\tiny{$\pm$59.19} & 292.68\tiny{$\pm$1.37}  & 260.42\tiny{$\pm$1.45}  & 209.45\tiny{$\pm$3.22}  \\
\textit{2048} & 290.82\tiny{$\pm$1.19}  & 254.81\tiny{$\pm$0.81}  & 197.16\tiny{$\pm$0.88}  & 161.64\tiny{$\pm$1.42}  \\
\textit{4096} & 251.34\tiny{$\pm$1.4}   & 187.19\tiny{$\pm$1.24}  & 154.32\tiny{$\pm$0.7}   & 148.07\tiny{$\pm$0.52}  \\
\textit{8192} & 181.0\tiny{$\pm$1.74}   & 149.79\tiny{$\pm$0.68}  & 145.92\tiny{$\pm$0.26}  & 145.55\tiny{$\pm$0.25}  \\ \bottomrule
\label{tab:lin_lba_census}

\end{tabular}
}
\subcaption{US Census Data}

    \end{minipage}
    \label{tab:lin_lba_error}
\end{table}

\begin{table}[!htb]
    \caption{Test MSE on training Neural Network on data created using \LLP with $\rho=0.1$ with error bars}
    \begin{minipage}{.5\linewidth}
        \centering
\resizebox{0.95\linewidth}{!}{
\begin{tabular}{lllll} \toprule
\multicolumn{5}{c}{Instance Level MSE = 174.55\tiny{$\pm$2.03}} \\ \hline
\multicolumn{1}{c|}{} & \multicolumn{4}{c}{\textbf{k}}                          \\
\multicolumn{1}{c|}{\textbf{m}}             & \textit{32}  & \textit{64}  & \textit{128} & \textit{256} \\ \toprule
\textit{512}  & 235.0\tiny{$\pm$25.35}  & 240.24\tiny{$\pm$32.01} & 229.65\tiny{$\pm$18.72} & 226.2\tiny{$\pm$12.74}  \\
\textit{1024} & 224.98\tiny{$\pm$22.92} & 211.36\tiny{$\pm$9.53}  & 212.91\tiny{$\pm$10.33} & 222.91\tiny{$\pm$35.3}  \\
\textit{2048} & 214.52\tiny{$\pm$19.12} & 206.42\tiny{$\pm$14.11} & 203.62\tiny{$\pm$6.98}  & 209.92\tiny{$\pm$13.59} \\
\textit{4096} & 201.96\tiny{$\pm$9.17}  & 200.81\tiny{$\pm$10.71} & 204.08\tiny{$\pm$15.75} & 205.18\tiny{$\pm$14.92} \\ \bottomrule
\label{tab:noisy_llp_criteo_0.1}

\end{tabular}
}
\subcaption{Criteo Sponsored Search Data}

    \end{minipage}%
    \begin{minipage}{.5\linewidth}
        \centering
\resizebox{0.95\linewidth}{!}{
\begin{tabular}{lllll} \toprule
\multicolumn{5}{c}{Instance Level MSE = 134.15\tiny{$\pm$0.12}} \\ \hline
\multicolumn{1}{c|}{} & \multicolumn{4}{c}{\textbf{k}}                          \\
\multicolumn{1}{c|}{\textbf{m}}             & \textit{64} & \textit{128} & \textit{256} & \textit{512} \\ \toprule
\textit{1024} & 156.64\tiny{$\pm$1.82} & 158.62\tiny{$\pm$3.49}  & 156.83\tiny{$\pm$2.72}  & 156.24\tiny{$\pm$2.53}  \\
\textit{2048} & 153.15\tiny{$\pm$1.98} & 151.94\tiny{$\pm$1.57}  & 152.93\tiny{$\pm$2.67}  & 155.5\tiny{$\pm$8.57}   \\
\textit{4096} & 149.15\tiny{$\pm$1.83} & 150.0\tiny{$\pm$2.85}   & 150.47\tiny{$\pm$3.71}  & 149.83\tiny{$\pm$3.33}  \\
\textit{8192} & 146.89\tiny{$\pm$1.49} & 146.45\tiny{$\pm$1.66}  & 145.9\tiny{$\pm$1.36}   & 148.48\tiny{$\pm$3.73} \\ \bottomrule
\label{tab:noisy_llp_census_0.1}

\end{tabular}
}
\subcaption{US Census Data}

    \end{minipage} 
    \label{tab:noisy_llp_error}
\end{table}

\begin{table}[!htb]
    \caption{Test MSE on training Neural Network on data created using \LLP with $\rho=0.0$}
    \begin{minipage}{.5\linewidth}
        \centering
\resizebox{0.95\linewidth}{!}{
\begin{tabular}{lllll} \toprule
\multicolumn{5}{c}{Instance Level MSE = 174.62\tiny{$\pm$2.14}} \\ \hline
\multicolumn{1}{c|}{} & \multicolumn{4}{c}{\textbf{k}}                          \\
\multicolumn{1}{c|}{\textbf{m}}             & \textit{32}  & \textit{64}  & \textit{128} & \textit{256} \\ \toprule
\textit{512}  & 241.14\tiny{$\pm$32.56} & 222.19\tiny{$\pm$10.54} & 231.13\tiny{$\pm$22.67} & 241.69\tiny{$\pm$24.45} \\
\textit{1024} & 211.58\tiny{$\pm$9.45}  & 217.82\tiny{$\pm$17.51} & 211.51\tiny{$\pm$14.63} & 214.24\tiny{$\pm$9.62}  \\
\textit{2048} & 207.32\tiny{$\pm$11.51} & 207.25\tiny{$\pm$12.75} & 201.09\tiny{$\pm$4.91}  & 209.23\tiny{$\pm$16.02} \\
\textit{4096} & 204.66\tiny{$\pm$17.62} & 197.77\tiny{$\pm$5.01}  & 197.31\tiny{$\pm$6.01}  & 202.67\tiny{$\pm$9.13} \\ \bottomrule
\label{tab:noisy_llp_criteo_gamma_0.0}

\end{tabular}
}
\subcaption{Criteo Sponsored Search Data}

    \end{minipage}%
    \begin{minipage}{.5\linewidth}
        \centering
\resizebox{0.95\linewidth}{!}{
\begin{tabular}{lllll} \toprule
\multicolumn{5}{c}{Instance Level MSE = 134.14\tiny{$\pm$0.12}} \\ \hline
\multicolumn{1}{c|}{} & \multicolumn{4}{c}{\textbf{k}}                          \\
\multicolumn{1}{c|}{\textbf{m}}             & \textit{64} & \textit{128} & \textit{256} & \textit{512} \\ \toprule
\textit{1024} & 157.05\tiny{$\pm$1.62} & 156.55\tiny{$\pm$2.19}  & 156.63\tiny{$\pm$2.75}  & 155.55\tiny{$\pm$2.78}  \\
\textit{2048} & 152.63\tiny{$\pm$2.28} & 152.38\tiny{$\pm$2.18}  & 151.72\tiny{$\pm$1.65}  & 152.99\tiny{$\pm$3.16}  \\
\textit{4096} & 149.46\tiny{$\pm$2.07} & 148.2\tiny{$\pm$1.28}   & 149.67\tiny{$\pm$2.98}  & 149.81\tiny{$\pm$2.86}  \\
\textit{8192} & 146.7\tiny{$\pm$1.04}  & 146.75\tiny{$\pm$2.49}  & 148.45\tiny{$\pm$3.81}  & 146.29\tiny{$\pm$1.54}  \\ \bottomrule
\label{tab:noisy_llp_census_gamma_0.0}

\end{tabular}
}
\subcaption{US Census Data}

    \end{minipage} 
    \label{tab:noisy_llp_gamma_0.0}
\end{table}

\begin{table}[!htb]
    \caption{Test MSE on training Neural Network on data created using \LLP with $\rho=0.01$}
    \begin{minipage}{.5\linewidth}
        \centering
\resizebox{0.95\linewidth}{!}{
\begin{tabular}{lllll} \toprule
\multicolumn{5}{c}{Instance Level MSE = 174.74\tiny{$\pm$2.19}} \\ \hline
\multicolumn{1}{c|}{} & \multicolumn{4}{c}{\textbf{k}}                          \\
\multicolumn{1}{c|}{\textbf{m}}             & \textit{32}  & \textit{64}  & \textit{128} & \textit{256} \\ \toprule
\textit{512}  & 227.66\tiny{$\pm$17.82} & 220.44\tiny{$\pm$10.09} & 224.3\tiny{$\pm$14.06}  & 240.18\tiny{$\pm$41.3}  \\
\textit{1024} & 211.51\tiny{$\pm$11.74} & 216.17\tiny{$\pm$11.6}  & 211.68\tiny{$\pm$8.95}  & 216.34\tiny{$\pm$8.64}  \\
\textit{2048} & 210.57\tiny{$\pm$16.11} & 215.05\tiny{$\pm$27.44} & 217.26\tiny{$\pm$21.04} & 206.18\tiny{$\pm$9.58}  \\
\textit{4096} & 198.7\tiny{$\pm$4.45}   & 206.05\tiny{$\pm$12.99} & 198.07\tiny{$\pm$6.55}  & 202.09\tiny{$\pm$12.16} \\ \bottomrule
\label{tab:noisy_llp_criteo_gamma_0.01}

\end{tabular}
}
\subcaption{Criteo Sponsored Search Data}

    \end{minipage}%
    \begin{minipage}{.5\linewidth}
        \centering
\resizebox{0.95\linewidth}{!}{
\begin{tabular}{lllll} \toprule
\multicolumn{5}{c}{Instance Level MSE = 134.14\tiny{$\pm$0.13}} \\ \hline
\multicolumn{1}{c|}{} & \multicolumn{4}{c}{\textbf{k}}                          \\
\multicolumn{1}{c|}{\textbf{m}}             & \textit{64} & \textit{128} & \textit{256} & \textit{512} \\ \toprule
\textit{1024} & 157.8\tiny{$\pm$1.69}  & 157.42\tiny{$\pm$3.4}   & 156.97\tiny{$\pm$2.5}   & 159.27\tiny{$\pm$9.4}   \\
\textit{2048} & 152.0\tiny{$\pm$1.21}  & 152.09\tiny{$\pm$1.31}  & 152.71\tiny{$\pm$3.07}  & 151.89\tiny{$\pm$1.95}  \\
\textit{4096} & 148.91\tiny{$\pm$0.99} & 148.89\tiny{$\pm$1.35}  & 148.3\tiny{$\pm$1.37}   & 152.7\tiny{$\pm$7.42}   \\
\textit{8192} & 147.14\tiny{$\pm$1.9}  & 146.78\tiny{$\pm$2.62}  & 146.1\tiny{$\pm$1.43}   & 146.28\tiny{$\pm$2.38}  \\ \bottomrule
\label{tab:noisy_llp_census_gamma_0.01}

\end{tabular}
}
\subcaption{US Census Data}

    \end{minipage} 
    \label{tab:noisy_llp_gamma_0.01}
\end{table}

\begin{table}[!htb]
    \caption{Test MSE on training Neural Network on data created using \LLP with $\rho=0.5$}
    \begin{minipage}{.5\linewidth}
        \centering
\resizebox{0.95\linewidth}{!}{
\begin{tabular}{lllll} \toprule
\multicolumn{5}{c}{Instance Level MSE = 174.53\tiny{$\pm$2.05}} \\ \hline
\multicolumn{1}{c|}{} & \multicolumn{4}{c}{\textbf{k}}                          \\
\multicolumn{1}{c|}{\textbf{m}}             & \textit{32} & \textit{64}  & \textit{128} & \textit{256} \\ \toprule
\textit{512}  & 223.15\tiny{$\pm$14.0} & 226.42\tiny{$\pm$17.11} & 227.86\tiny{$\pm$15.59} & 236.18\tiny{$\pm$19.4}  \\
\textit{1024} & 209.06\tiny{$\pm$3.38} & 221.71\tiny{$\pm$20.75} & 238.25\tiny{$\pm$30.97} & 221.74\tiny{$\pm$15.37} \\
\textit{2048} & 204.89\tiny{$\pm$7.72} & 215.81\tiny{$\pm$17.6}  & 213.26\tiny{$\pm$12.59} & 215.96\tiny{$\pm$35.53} \\
\textit{4096} & 200.57\tiny{$\pm$4.5}  & 202.94\tiny{$\pm$14.87} & 202.84\tiny{$\pm$9.67}  & 197.66\tiny{$\pm$4.56}  \\ \bottomrule
\label{tab:noisy_llp_criteo_gamma_0.5}

\end{tabular}
}
\subcaption{Criteo Sponsored Search Data}

    \end{minipage}%
    \begin{minipage}{.5\linewidth}
        \centering
\resizebox{0.95\linewidth}{!}{
\begin{tabular}{lllll} \toprule
\multicolumn{5}{c}{Instance Level MSE = 134.14\tiny{$\pm$0.11}} \\ \hline
\multicolumn{1}{c|}{} & \multicolumn{4}{c}{\textbf{k}}                          \\
\multicolumn{1}{c|}{\textbf{m}}             & \textit{64} & \textit{128} & \textit{256} & \textit{512} \\ \toprule
\textit{1024} & 157.23\tiny{$\pm$1.63} & 156.55\tiny{$\pm$1.56}  & 155.84\tiny{$\pm$2.54}  & 158.64\tiny{$\pm$6.35}  \\
\textit{2048} & 151.93\tiny{$\pm$1.17} & 151.54\tiny{$\pm$1.49}  & 153.22\tiny{$\pm$4.27}  & 152.45\tiny{$\pm$2.43}  \\
\textit{4096} & 148.28\tiny{$\pm$0.82} & 149.43\tiny{$\pm$1.97}  & 148.81\tiny{$\pm$2.03}  & 153.35\tiny{$\pm$10.78} \\
\textit{8192} & 146.5\tiny{$\pm$2.0}   & 146.42\tiny{$\pm$2.14}  & 146.81\tiny{$\pm$2.95}  & 147.45\tiny{$\pm$3.59}  \\ \bottomrule
\label{tab:noisy_llp_census_gamma_0.5}

\end{tabular}
}
\subcaption{US Census Data}

    \end{minipage} 
    \label{tab:noisy_llp_gamma_0.5}
\end{table}

\begin{table}[!htb]
    \caption{Test MSE on training Neural Network on data created using \LLP with $\rho=1.0$}
    \begin{minipage}{.5\linewidth}
        \centering
\resizebox{0.95\linewidth}{!}{
\begin{tabular}{lllll} \toprule
\multicolumn{5}{c}{Instance Level MSE = 174.35\tiny{$\pm$2.05}} \\ \hline
\multicolumn{1}{c|}{} & \multicolumn{4}{c}{\textbf{k}}                          \\
\multicolumn{1}{c|}{\textbf{m}}             & \textit{32}  & \textit{64} & \textit{128} & \textit{256} \\ \toprule
\textit{512}  & 225.1\tiny{$\pm$13.61}  & 221.0\tiny{$\pm$12.16} & 223.77\tiny{$\pm$10.71} & 229.39\tiny{$\pm$18.22}\\
\textit{1024} & 222.44\tiny{$\pm$26.83} & 212.7\tiny{$\pm$10.77} & 215.23\tiny{$\pm$15.55} & 211.46\tiny{$\pm$15.07} \\
\textit{2048} & 209.45\tiny{$\pm$13.0}  & 205.31\tiny{$\pm$10.5} & 206.08\tiny{$\pm$8.43}  & 214.62\tiny{$\pm$17.26} \\
\textit{4096} & 196.72\tiny{$\pm$3.8}   & 198.02\tiny{$\pm$4.81} & 205.25\tiny{$\pm$14.11} & 193.41\tiny{$\pm$2.84}  \\ \bottomrule
\label{tab:noisy_llp_criteo_gamma_1.0}

\end{tabular}
}
\subcaption{Criteo Sponsored Search Data}

    \end{minipage}%
    \begin{minipage}{.5\linewidth}
        \centering
\resizebox{0.95\linewidth}{!}{
\begin{tabular}{lllll} \toprule
\multicolumn{5}{c}{Instance Level MSE = 134.15\tiny{$\pm$0.11}} \\ \hline
\multicolumn{1}{c|}{} & \multicolumn{4}{c}{\textbf{k}}                          \\
\multicolumn{1}{c|}{\textbf{m}}             & \textit{64} & \textit{128} & \textit{256} & \textit{512} \\ \toprule
\textit{1024} & 156.87\tiny{$\pm$1.72} & 158.45\tiny{$\pm$2.47}  & 156.6\tiny{$\pm$2.1}    & 159.17\tiny{$\pm$4.58}  \\
\textit{2048} & 152.91\tiny{$\pm$2.23} & 151.91\tiny{$\pm$1.07}  & 151.53\tiny{$\pm$1.0}   & 152.84\tiny{$\pm$5.2}   \\
\textit{4096} & 148.26\tiny{$\pm$0.6}  & 149.0\tiny{$\pm$1.65}   & 150.15\tiny{$\pm$3.24}  & 151.75\tiny{$\pm$5.96}  \\
\textit{8192} & 146.31\tiny{$\pm$1.33} & 145.87\tiny{$\pm$1.9}   & 146.54\tiny{$\pm$1.49}  & 146.2\tiny{$\pm$0.93}   \\ \bottomrule
\label{tab:noisy_llp_census_gamma_1.0}

\end{tabular}
}
\subcaption{US Census Data}

    \end{minipage} 
    \label{tab:noisy_llp_gamma_1.0}
\end{table}

\end{document}